\documentclass[lettersize,journal]{IEEEtran}
\usepackage{amsmath,amsfonts}
\usepackage{array}
\usepackage[caption=false,font=normalsize,labelfont=sf,textfont=sf]{subfig}
\usepackage{textcomp}
\usepackage{xcolor}
\usepackage{stfloats}
\usepackage{url}
\usepackage{verbatim}
\usepackage{graphicx}
\usepackage{cite}
\usepackage{booktabs}
\usepackage{makecell}
\usepackage[table]{xcolor}
\usepackage[switch,mathlines]{lineno}
\usepackage[normalem]{ulem}
\usepackage{enumitem}

\usepackage{algorithm}
\usepackage{algpseudocode}
\usepackage[hidelinks]{hyperref}

\usepackage{amsmath,amssymb}

\usepackage{makecell}   
\usepackage{pifont} 
\newcommand{\cmark}{\ding{51}}   
\newcommand{\xmark}{\ding{55}}   

\newcommand{\umark}{\ensuremath{\uparrow}}   
\newcommand{\dmark}{\ensuremath{\downarrow}} 

\definecolor{meetbg}{HTML}{E8F5E9} 
\definecolor{failbg}{HTML}{FFEBEE} 
\newcommand{\meetrow}{\rowcolor{meetbg}}
\newcommand{\failrow}{\rowcolor{failbg}}

\begin{document}

\title{SigmaQuant: Hardware-Aware Heterogeneous Quantization Method for Edge DNN Inference}




\author{
	\IEEEauthorblockN{
	Qunyou Liu\IEEEauthorrefmark{1},
    Pengbo Yu\IEEEauthorrefmark{1},
    Marina Zapater\IEEEauthorrefmark{2},
    David Atienza\IEEEauthorrefmark{1},  \\
	}			  
    \IEEEauthorblockA{\IEEEauthorrefmark{1}\textit{Embedded Systems Laboratory (ESL), EPFL, Switzerland, }\textit{qunyou.liu@epfl.ch, pengbo.yu@epfl.ch, david.atienza@epfl.ch}}    
    \IEEEauthorblockA{\IEEEauthorrefmark{2}\textit{University of Applied Sciences and Arts Western Switzerland (HES-SO), Switzerland, }\textit{marina.zapater@heig-vd.ch}}
} 
\IEEEaftertitletext{\vspace{-2.5\baselineskip}} 

\maketitle

\begin{abstract}
Deep neural networks (DNNs) are essential for performing advanced tasks on edge or mobile devices, yet their deployment is often hindered by severe resource constraints, including limited memory, energy, and computational power. While uniform quantization provides a straightforward approach to compress model and reduce hardware requirement, it fails to fully leverage the varying robustness across layers, and often lead to accuracy degradation or suboptimal resource usage, particularly at low bitwidths. In contrast, heterogeneous quantization, which allocates different bitwidths to individual layers, can mitigate these drawbacks. Nonetheless, current heterogeneous quantization methods either needs huge brute-force design space search or lacks the adaptability to meet different hardware conditions, such as memory size, energy budget, and latency requirement. Filling these gaps, this work introduces \textbf{\textit{SigmaQuant}}, an adaptive layer-wise heterogeneous quantization framework designed to efficiently balance accuracy and resource usage for varied edge environments without exhaustive search.
SigmaQuant allocates layer-wise bitwidths based on weight standard deviation and KL divergence, enabling adaptive quantization under real hardware constraints. This strategy efficiently balances accuracy, memory, and latency without the use of exhaustive search. We validate its practicality through ASIC integration in a shift-add-based accelerator, analyzing power, performance, and area (PPA) trade-offs for the resulting mixed-precision models.
Experimentals on CIFAR-100 and ImageNet show that SigmaQuant consistently outperforms both uniform and state-of-the-art heterogeneous quantization. At an equal size model, it achieves up to 2.0\% higher accuracy; at an equal accuracy, it reduces memory by up to 40.0\%. Hardware evaluation demonstrates up to 22.3\% area savings and 20.6\% less energy cost compared to the widely used INT8 quantization and implementation, with slight latency overhead and comparable accuracy. These results confirm the effectiveness of SigmaQuant for edge AI deployment.


\end{abstract}


\section{Introduction}
\label{sec:intro}

Deep Neural Networks (DNNs) have become the backbone of many edge computing applications, ranging from image processing on devices~\cite{10.1145/3065386} to speech recognition~\cite{10.5555/3045390.3045410} and beyond. However, executing these models with strict resource budgets remains a formidable challenge. In particular, edge devices often have extremely limited memory size~\cite{Liu2025Matrix,Liu2025Gem5AcceSys}, processing capabilities, and stringent energy supplies, which can make large and high-precision DNNs prohibitively expensive or even infeasible to deploy~\cite{Sze2017Efficient,Deng2020ModelCompressSurvey}.

Quantization has emerged as one of the most effective strategies to tackle this problem, as it converts 32-bit floating point (FP) weights and activations into lower-precision representations (e.g., 8-bit or even 4-bit integers). This compression reduces the storage footprint and computational overhead of the model \cite{Han2016DeepCompression,Jacob2018Quantization}. However, \emph{uniform} quantization, where all layers share the same bitwidth, often fails to achieve an optimal trade-off between accuracy and efficiency. The reason is that each DNN layer exhibits distinct statistical properties and different robustness to quantization noise \cite{Krishnamoorthi2018QuantTutorial,banner2018scalable}. Therefore, forcing a single global precision can over‐allocate bits to some layers and under‐allocate them to others.

To overcome these limitations, \emph{heterogeneous} quantization has recently gained popularity. This approach tailors the bitwidth per layer based on the quantization sensitivity of each layer, resulting in more compact models with minimal accuracy loss. However, existing methods typically rely on layer-wise sensitivity heuristics or resource-intensive search algorithms (e.g., reinforcement learning)~\cite{yao2021hawq,Baskin2021uniq}. In addition, edge devices exhibit various hardware configurations and resource constraints. For example, IoT sensors might have only a few megabytes of available main memory, whereas mobile phones can offer significantly more but require higher accuracy for applications like real-time image recognition or speech processing. Consequently, a static quantization scheme optimized for one scenario often fails to generalize to other hardware deployments. Formally, the quantization process should ideally satisfy \textit{boundary conditions} on memory and accuracy constraints:
\[
\begin{aligned}
\text{Memory Usage} &\le \text{Memory Constraint},\\[2pt]
\text{Accuracy}     &\ge \text{Accuracy Constraint}.
\end{aligned}
\]

Current heterogeneous quantization methods fail to satisfy this adaptive requirement, as they typically provide fixed solutions that cannot accommodate the diverse and dynamic demands of edge scenarios.
Beyond algorithmic efficiency, heterogeneous quantization must also deliver tangible hardware benefits, particularly when deployed on edge AI accelerators. These accelerators increasingly rely on reduced-precision arithmetic to efficiently minimize area, power, and latency. The widely adopted shift-add scheme for multiply-accumulate (MAC) operations constitutes a representative example~\cite{judd2016stripes, hsu2020essa, rios2023bit, yu2024energy}. In such cases, lower bitwidths significantly accelerate MAC computation, as fewer bits directly translate into less processing cycles and lower energy consumption. However, simply applying the lowest precision everywhere is ineffective, since some layers are highly sensitive to quantization noise. Over-quantizing these layers can lead to severe accuracy loss, undermining any hardware gains~\cite{dong2019hawq,yazdanbakhsh2019releq}. This motivates a hardware-aware heterogeneous quantization strategy that carefully allocates precision across the network—assigning low bitwidths where feasible and preserving higher precision where needed. Such an approach ensures compliance with hardware  constraints (e.g., memory size, energy budgets, real-time latency) and maximizes the hardware efficiency.

For this purpose, we propose SigmaQuant, a hardware-aware heterogeneous quantization framework that adaptively allocates per-layer bitwidth under user-specified memory and accuracy targets, while fully leveraging hardware efficiency simultaneously. SigmaQuant has two search phases: (i) Phase-1 provides a stable, constraint-aware initialization. All layers are grouped by simple distribution statistics so that the model quickly moves into a region that satisfies at least one boundary (memory size or accuracy). (ii) Phase-2 then performs fine-grained adjustments on a few layers each time to meet both targets while avoiding unnecessary changes in most layers. The proposed two-phase scheme reduces the search overhead and effectively satisfies the requirements imposed by tight resource budgets. We employ \textit{SigmaQuant} on CIFAR-100 and ImageNet datasets and on a set of DNNs (e.g., ResNet and MobileNet)  and compare performance with respect to a uniform quantization scheme and several state-of-the-art heterogeneous quantization techniques. Furthermore, we evaluate \textit{SigmaQuant} on a general shift-add-based MAC unit widely used in modern edge accelerators, demonstrating its effectiveness in reducing latency and energy. We also provide a detailed trade-off analysis that highlights the co-optimization between software quantization strategies and hardware design, showing how SigmaQuant impacts performance, power, and area (PPA) metrics. This underscores the importance of proposing quantization methods that simultaneously optimize algorithmic performance while adhering to hardware constraints for efficient deployment.
More specifically, the \textbf{key contributions} of our work are the following:
\begin{itemize}[leftmargin=*,labelsep=0.4em,itemsep=0.2em,topsep=0.2em]
\item  We introduce a \textbf{distribution-based approach} to study heterogeneous quantization, fully utilizing the weight of each layer, guided by standard deviation and KL divergence.

\item We develop a \textbf{two-phase quantization method} that combines a cluster-based assignment with iterative divergence-driven refinement, striking an effective accuracy--efficiency balance in a reasonable design space search. Our method adapts the layer bitwidth assignments according to the user-defined \emph{boundary conditions} on accuracy and model size, ensuring minimal quantization loss while aggressively compressing the network.

\item We \textbf{validate our method} on CIFAR-100~\cite{Krizhevsky2009Learning} and ImageNet~\cite{Russakovsky2015ImageNet} with popular DNN architectures (mainly ResNet and MobileNet families). Compared to state-of-the-art heterogeneous schemes, our approach (1) reduces memory usage by up to \textbf{17.7\%} while maintaining or exceeding accuracy and (2) reaches over 2\% in Top-1 accuracy under similar memory budgets. Compared to uniform quantization, \textit{SigmaQuant} can reach the same accuracy with only 60\% memory budget, and achieve 4\% more accuracy with the same model size.

\item We evaluate \textbf{SigmaQuant} using a generic hardware implementation based on the shift-add scheme, which is widely used in edge scenarios. The results show that our method provides plenty of quantization choices between latency/energy and accuracy. It outperforms uniform quantization schemes such as A8W4 in all metrics. Compared to a well-performing INT8 quantization strategy and hardware, \textbf{SigmaQuant} can further reduce area by 22.3\% and energy consumption by up to 20. 6\% with slight latency overhead and comparable accuracy.
\end{itemize}

The remainder of this paper is organized as follows. Section~\ref{sec:related_work} describes the most relevant heterogeneous quantization research in the state of the art. Section~\ref{sec:background_distribution} explains the mathematical explanation for the algorithm and hardware background of suitable accelerator used in performance validation. Section~\ref{sec:methodology} details our distribution-fitting approach and two-phase algorithm. Section~\ref{sec:experiments} describes our experimental setup, followed by results and analysis in Section~\ref{sec:results}. We conclude and discuss future directions in Section~\ref{sec:conclusion}.

\section{Related Work}
\label{sec:related_work}
Deep neural network quantization has seen extensive research to balance accuracy and efficiency for deployment on resource-constrained hardware. A common approach is uniform quantization, where all weights and activations are quantized to a fixed bitwidth using a linear mapping scale. Uniform low-precision (e.g., 8-bit integer) models can significantly reduce memory usage and speed up inference with only minor accuracy loss in many cases~\cite{yao2021hawq}. However, pushing uniform quantization to ultra-low bitwidths (4-bit or below) often incurs severe accuracy degradation~\cite{yao2021hawq}. Techniques such as quantization-aware training and careful calibration improve the situation~\cite{yao2021hawq}, but ultimately a one-size-fits-all bitwidth may not suit all layers. Some layers are more sensitive to quantization errors than others (as described in Section~\ref{subsec:distribution_fitting}), so using the same precision budget for every layer can be suboptimal (as shown in Section~\ref{subsec:modelsize_accuracy_cifar100}). This rigidity motivates the use of adaptive quantization strategies.

Non-uniform quantization techniques allocate quantization levels unevenly, guided by the sensitivity of layers. Rather than spacing quantization levels uniformly, these methods concentrate precision where it matters the most (e.g., large-magnitude values). For instance, clustering or entropy-based schemes learn value clusters that maximize information content~\cite{8100244}, effectively compressing many small-magnitude weights into fewer levels while reserving more levels for important outliers~\cite{8100244}. Zhu et al. propose an entropy-aware layerwise quantization where the bitwidth of each layer is chosen based on the weight distribution entropy, allowing more complex layers to use higher precision (and vice versa)~\cite{ZhuEntropy2018}. Baskin et al.~\cite{Baskin2021uniq} introduce a non-uniform $k$-quantile approach (UNIQ) that adapts the quantizer to the parameter distribution by injecting noise during training to emulate a quantile-based quantization. These non-uniform quantizers can better preserve accuracy under aggressive compression, especially in the low-bit regime~\cite{Baskin2021uniq}, but their irregular quantization levels offer only a limited set of solutions and may not reliably meet memory constraints.

Recent approaches formulate mixed-precision selection as an optimization problem under hardware constraints. Hardware-Aware Quantization frameworks like HAQ~\cite{8954415} employ reinforcement learning to search for the Pareto-optimal bit allocation across layers given a target memory or latency budget, automating mixed‑precision selection. Its runtime depends on the training budget and on the details of the implementation. Likewise, differentiable Neural Architecture Search~\cite{10.5555/3322706.3361996} techniques have been applied to quantization, learning per-layer precisions via gradient-based optimization \cite{Wu2018FBNetHE}, but this approach entails training a large super-network (for bitwidth choices) – incurring considerable computational overhead – and can be sensitive to initialization and hyper-parameters. 
Second-order sensitivity analysis is also well established in statistics and machine learning~\cite{cai2021tsp,NEURIPS2020_d77c7035}. An alternative line of work uses second-order sensitivity (HAWQ -- Hessian-AWare Quantization). HAWQ leverages the Hessian spectrum of each layer to gauge how quantization will impact the loss~\cite{NEURIPS2020_d77c7035}. Heterogeneous DNNs have demonstrated that judiciously allocating higher bitwidth to error-sensitive layers and lower precision to tolerant layers can reduce model size and inference cost with only a small accuracy drop. 

There are also approaches for dynamic or adaptive precision at runtime. Instead of fixing a mixed-precision configuration post-training, a single model is trained to support multiple bitwidth settings. Jin et al.~\cite{Adabits2020} propose AdaBits, which enables a network to switch between bitwidths on the fly (e.g., 2-bit to 8-bit) without retraining. By joint training with adaptive bitwidths and using techniques like switchable clipping levels for activations, they obtain a model that can cater to different hardware capabilities or energy budgets. This adds a new dimension of flexibility: one deployed model can adapt to various scenarios. The trade-off is a more involved training procedure (to ensure the network performs well at all supported precisions) and a slight runtime overhead to manage precision switching.

Existing network quantization methods, including uniform and mixed-precision approaches, often do not explicitly minimize the divergence between the pre- and post-quantization distributions of weights or activations. Instead, they use indirect criteria (e.g. minimizing quantization noise variance or heuristics like entropy/KL to set ranges), which do not guarantee that the quantized values preserve the original distribution~\cite{Finkelstein2019FightingQB}. This oversight can lead to a mismatch in distributions that degrades accuracy, as recently noted by several studies~\cite{Finkelstein2019FightingQB}~\cite{hong2024overcoming}. Moreover, most quantization frameworks produce a fixed precision configuration for a model. Many mixed-precision techniques explore only a handful of bitwidth assignments and then adopt a static scheme, which lacks flexibility across different hardware settings~\cite{9709904}. This nature means the model cannot easily adapt its bitwidth per layer to meet the constraints or capabilities of new devices. In summary, earlier quantization methods did not explicitly align distributions and often yielded fixed precision setups.

While heterogeneous quantization effectively tailors precision to each layer’s tolerance, deployment on real hardware demands further consideration. Edge AI accelerators, whether ASIC, DSIP, DSP, FPGA, etc., operate under strict area, power, and latency constraints. Reduced-precision arithmetic directly affects all these metrics, as lowering bitwidth reduces MAC complexity, diminishes the memory footprint and data movement, and thereby decreases power consumption. For example, for the shift-add scheme widely used in many accelerators, each additional bit in a multiplier operand contributes an extra cycle and corresponding power cost. Thus, lowering the weight bitwidth yields an approximately proportional reduction in compute latency and energy consumption per operation. However, naively pushing all layers to the lowest precision can devastate accuracy, since sensitive layers quickly lose accuracy under over-quantization. Existing quantization techniques largely neglect this co-optimization. Many prior methods focus on accuracy and model size (often via heuristic or search-based bit allocations) without explicitly modeling hardware costs. These approaches yield fixed precision assignments that may satisfy a bit-budget or target accuracy but run afoul of practical limits like on-chip memory capacity, power budgets, or throughput constraints. In other words, a model might be well-compressed in theory but still too slow or energy-hungry for the target hardware. This gap in prior work motivates a hardware-aware solution that jointly addresses algorithmic and hardware objectives.
Beyond early mixed-precision schemes such as HAQ and HAWQ, recent work has progressed along several complementary directions that explicitly consider hardware constraints and/or more expressive search formulations. 
On the hardware--software co-design side, Edge-MPQ~\cite{Zhao2024EdgeMPQ} integrates versatile mixed-precision inference units into a RISC-V pipeline and couples them with a hardware-aware layer-wise MPQ search, targeting latency and energy on a specific edge platform. 
On the training/sensitivity side, Huang et al.~\cite{Huang2025HMQAT} propose HMQAT, a Hessian-based mixed-precision QAT framework that drives bit configuration via a Pareto-frontier search and further improves stability through transition-aware fine-tuning of quantization scales. 
From the perspective of joint compression under an accelerator objective, Balaskas et al.~\cite{Balaskas2024DiversePruning} explore a combined pruning--quantization design space with a reinforcement-learning agent guided by an accelerator-level energy model. 
Finally, Deng et al.~\cite{Deng2025CLADO} present CLADO, which explicitly models cross-layer dependencies of quantization errors and formulates bitwidth assignment as an integer quadratic program to obtain strong accuracy--compression trade-offs.

Motivated by these advances and their practical trade-offs, our proposed approach, \emph{SigmaQuant}, adopts a lightweight yet hardware-aware distribution-fitting perspective for heterogeneous quantization. Specifically, SigmaQuant uses layer-wise standard deviation and the KL divergence between the floating-point and quantized distributions to guide bitwidth assignment in a two-phase procedure that enforces user-defined accuracy and resource constraints (model size or BOPs) without exhaustive search, while remaining compatible with integer-only shift-add accelerators.

\section{Background}
\label{sec:background_distribution}

In this section, we first review the conventional (uniform) quantization scheme and then introduce our perspective of approximating each layer’s weight distribution by a discrete (quantized) distribution. This \emph{distribution-fitting} viewpoint motivates why different layers may require distinct bitwidths. We also highlight the increasing importance of hardware-constrained design, wherein bitwidth choices must align with metrics such as model size to meet resource and latency targets on edge devices, and we provide some background on the specific hardware design that we use to showcase the results of our work.

\subsection{Model Quantization and Distribution}
\label{subsec:distribution_fitting}
\subsubsection{Uniform Quantization Basics.}
Let $\mathbf{w} \in \mathbb{R}^n$ denote the set of weights in a given DNN layer. In a uniform quantization scheme with integer bitwidth $b$, each weight $w_i$ is replaced by a quantized value 
\[
    \tilde{w}_i \;=\; \mathrm{clip}\!\Big(\mathrm{round}\!\big(\frac{w_i}{\Delta}\big),\, -Q,\, Q\Big)\times \Delta,
\] 
where $Q = 2^{(b-1)} - 1$ (for signed quantization) and $\Delta$ is the quantization step size (or scale).  Typically, $\Delta$ is chosen in one of two ways: 
\begin{itemize}[leftmargin=*,labelsep=0.4em,itemsep=0.2em,topsep=0.2em]\itemsep0pt
    \item \textit{Max-based scaling}, where $\displaystyle \Delta = \max_i |w_i| / Q$,
    \item \textit{Statistical scaling}, where $\displaystyle \Delta = k\,\sigma$, with $\sigma$ denoting the standard deviation of $\mathbf{w}$ (and $k$ a chosen constant, e.g., $2$ or $3$).
\end{itemize} 
All weights are thus mapped to one of $2^b$ uniformly spaced levels in $[-Q\Delta,\; Q\Delta]$. While straightforward, a global uniform bitwidth can be suboptimal because different layers exhibit widely varying sensitivities to quantization noise~\cite{Baskin2021uniq,yao2021hawq,NEURIPS2020_d77c7035,mishra2018apprentice}. This observation motivates \emph{heterogeneous quantization}, wherein each layer (or channel) is allowed its own bitwidth $b_\ell$. Such layer-wise customization yields better accuracy–size trade-offs: for example, early convolutional layers often have smaller weight variance and can safely be quantized to 4 bits, whereas layers with broader weight distributions may require 8 bits to retain accuracy. The challenge, then, is to determine how to assign precision across layers in a principled yet efficient way.

\subsubsection{Quantization Sensitivity and Standard Deviation}
We conducted preliminary experiments to identify a simple indicator of layer-wise quantization sensitivity. Using ImageNet~\cite{Russakovsky2015ImageNet} as dataset and Alexnet~\cite{alexnet} and ResNet18~\cite{resnet} as example models, we applied progressively lower bitwidths per layer to reach minimum Bits Operations (BOPs)~\cite{Baskin2021uniq} under a fixed overall precision budget using brute-force search. A remarkably strong correlation emerged between each layer’s acceptable bitwidth and the standard deviation $\sigma$ (\textit{Sigma}) of its weights: layers with relatively low $\sigma$ tolerated very low precision (4-bit or even 2-bit) with minimal accuracy impact, whereas layers with high $\sigma$ either needed to remain at higher precision (e.g., 8-bit) or suffered severe accuracy drops when forced to fewer bits. Table~\ref{tab:alexnet-sigma} illustrates this trend for AlexNet running on ImageNet, where an initial heuristic assignment (uniform 8-bit for all layers) is compared against the final adjusted bitwidths. Layers with larger $\sigma$ (e.g., \texttt{Conv1}) ended up requiring higher $b_\ell$ to avoid excessive quantization error, as reflected by the larger KL divergence $D_{KL}$ between the original and quantized weight distributions for those layers. In contrast, layers with small $\sigma$ (e.g., the fully-connected layers) were quantized to 2 bits with negligible $D_{KL}$ penalty. ResNet18 shows the same trend.

\begin{table}[h]
  \centering
  \caption{Observations on a baseline (BOP-based) heuristic vs. final bitwidth and weight distribution for AlexNet (ImageNet). Higher standard deviation $\sigma$ correlates with the need for higher bitwidth to keep the KL divergence $D_{KL}$ low.}
  \label{tab:alexnet-sigma}
  \begin{tabular}{lcccc}
    \toprule
    \textbf{Layer} & \textbf{Init Bits} & \textbf{Final Bits} & $\boldsymbol{\sigma}$ & $D_{KL}$ \\
    \midrule
    AlexNet -- Conv1 & 8 & 6 & 0.115672 & 0.022229 \\
    AlexNet -- Conv2 & 8 & 6 & 0.046543 & 0.000626 \\
    AlexNet -- Conv3 & 8 & 4 & 0.034646 & 0.000477 \\
    AlexNet -- Conv4 & 8 & 4 & 0.027320 & 0.000409 \\
    AlexNet -- Conv5 & 8 & 4 & 0.026128 & 0.000619 \\
    AlexNet -- FC1   & 8 & 2 & 0.009245 & 0.000097 \\
    AlexNet -- FC2   & 8 & 2 & 0.011537 & 0.000104 \\
    AlexNet -- FC3   & 8 & 4 & 0.018524 & 0.000136 \\
    \bottomrule
  \end{tabular}
\end{table}

These empirical findings confirm that the weight standard deviation $\sigma_\ell$ is a strong first-order indicator of layer $\ell$’s quantization sensitivity. Intuitively, $\sigma_\ell$ gauges the “width” of the weight distribution: a large spread means more information can be lost when mapping to a small set of levels, whereas a tightly concentrated distribution can be compressed more aggressively. This insight aligns with prior observations that layer-wise weight distributions influence quantization outcomes. For instance, the quantization whitepaper in~\cite{Krishnamoorthi2018QuantTutorial} recommends setting the quantizer range based on multiples of the weight standard deviation, and Zhu \textit{et al.} propose an entropy-based bit allocation per layer to account for distribution complexity~\cite{ZhuEntropy2018}. More sophisticated frameworks have also quantified layer sensitivity via reinforcement learning or second-order analysis (Hessian)~\cite{yao2021hawq}, reinforcing the notion that certain layers are inherently more error-prone under quantization. Our results show that a simple statistic $\sigma_\ell$ can serve as an effective proxy for this sensitivity, providing a convenient starting point for assigning heterogeneous bitwidths. In practice, this means we can sort or cluster layers by $\sigma_\ell$ and assign lower precisions to those with small $\sigma_\ell$ while reserving higher bits for layers with large $\sigma_\ell$, before applying any fine-grained adjustments.

\subsubsection{Distribution-Fitting Perspective.}
While $\sigma$ provides a useful heuristic, we also adopt a theoretical perspective to quantify quantization loss more rigorously. Consider the empirical distribution of the weights in layer $\ell$, denoted $p_\ell(w)$. We can write 
\[ 
    p_\ell(w) \;=\; \frac{1}{|W_\ell|}\sum_{w_i \in W_\ell} \delta(w - w_i)\,,
\] 
where $W_\ell=\{w_i\}$ is the set of weights and $\delta(\cdot)$ is the Dirac delta (this can be thought of as the normalized weight histogram for layer $\ell$). After quantizing layer $\ell$ to $b_\ell$ bits, the set of quantized weights $\{\tilde{w}_i\}$ induces a discrete distribution $\tilde{p}_\ell(w)$ supported only on the finite quantization levels. Rather than viewing quantization merely as per-weight rounding error, we regard it as an \emph{information-theoretic approximation} of $p_\ell$ by $\tilde{p}_\ell$. A natural measure of the mismatch between these two distributions is the Kullback–Leibler (KL) divergence:
\begin{equation}
   D_{KL}\!\big(p_\ell \parallel \tilde{p}_\ell\big) \;=\; \sum_{w}\; p_\ell(w)\;\log \frac{p_\ell(w)}{\tilde{p}_\ell(w)}\,.
   \label{eq:kl}
\end{equation}

When the quantization step size (represented by \( b_l \)) is too small compared to the range of values in \( W_l \), the quantized representation \( \tilde{p_l} \) will not capture the differences between many distinct values, leading to a loss of precision. Conversely, a sufficiently high $b_\ell$ yields $\tilde{p}_\ell$ that closely approximates $p_\ell$, resulting in a small KL divergence. In essence, choosing an appropriate bitwidth $b_\ell$ amounts to balancing model size (or efficiency) against an acceptable level of distribution distortion. This distribution-fitting view provides a theoretical foundation for quantization: one seeks to minimize information loss by ensuring $\tilde{p}_\ell$ is a good fit to $p_\ell$.

Importantly, this perspective helps explain the earlier empirical trend with $\sigma_\ell$. A layer with a large $\sigma_\ell$ typically has a broad $p_\ell(w)$; quantizing it aggressively (small $b_\ell$) forces a poor approximation $\tilde{p}_\ell(w)$, yielding a high $D_{KL}$ and, ultimately, higher accuracy degradation. In contrast, a layer with small $\sigma_\ell$ (narrow distribution) can be quantized to low precision with only a minor increase in $D_{KL}$. In practice, we leverage $D_{KL}$ as a quantification of quantization “distortion” to complement the $\sigma_\ell$ heuristic. Recent studies similarly note that preserving the original distribution (or minimizing distribution shift) is critical for maintaining accuracy~\cite{10.5555/3540261.3541610}. Thus, by using $\sigma_\ell$ as a first-order guide and $D_{KL}(p_\ell \parallel \tilde{p}_\ell)$ as a refinement criterion, we ground our heterogeneous quantization strategy in both empirical intuition and theoretical approximation. Subsequent sections will reference these metrics as the basis for our two-phase bitwidth assignment algorithm, without needing to repeat the justification for why they are chosen.

\subsection{Exploiting SigmaQuant in Efficient shift-add-based Accelerators for Edge Devices}
\label{sec:hardware_design}

\begin{figure}[ht]
    \centering
    \includegraphics[width=1.0\linewidth]{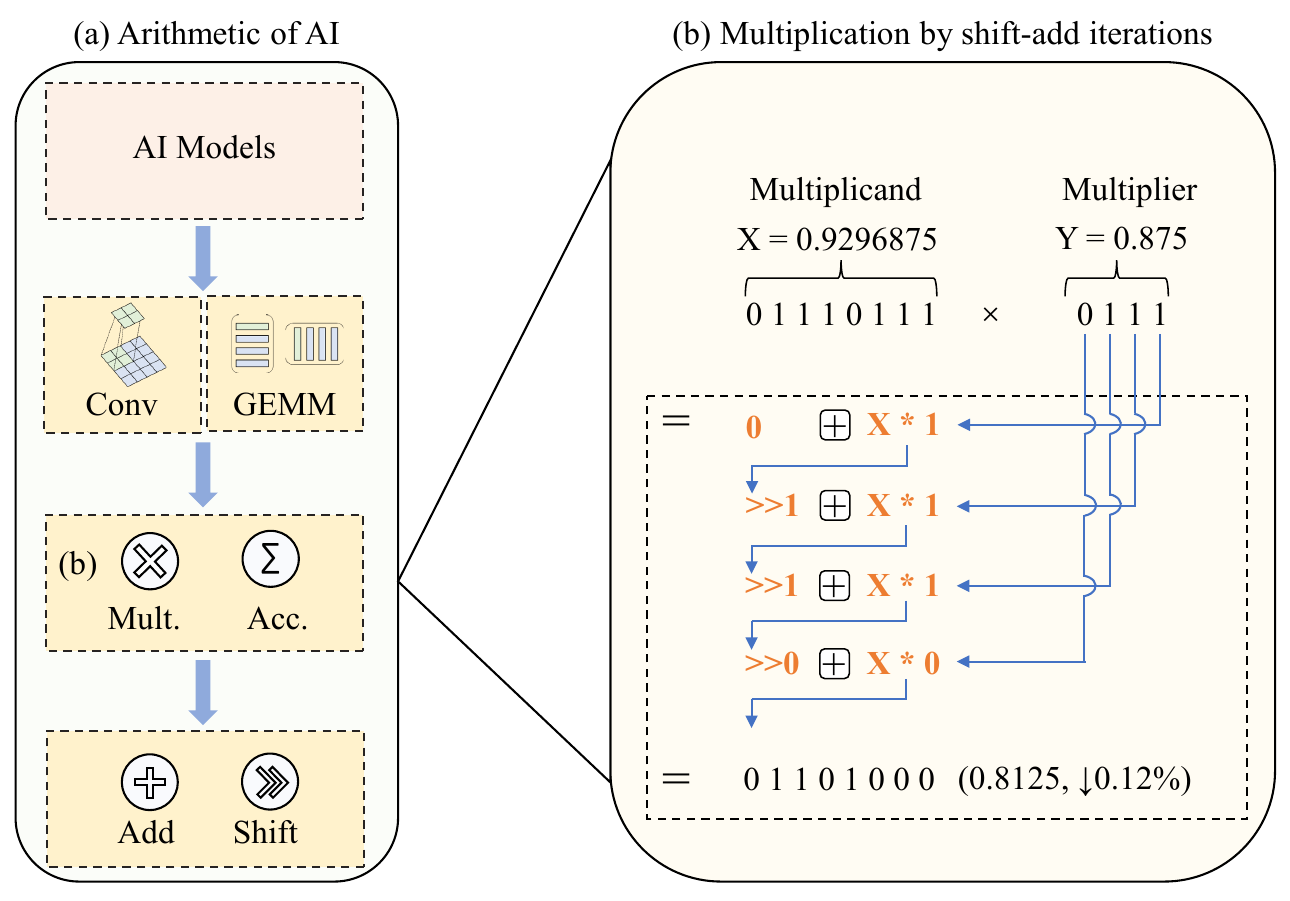}
    \caption{(a) The operation hierarchy of AI models from top to down, and (b) the widely used shift-add-based multiplier for edge accelerators that target at high energy efficiency. }
    \label{fig:hardware_diagram}
\end{figure}

Modern edge and mobile devices are constrained by limited memory, computational power, and energy resources, which significantly hampers DNN inference on these platforms. To increase the speed-up and efficiency of inference in resource-constrained platforms, models are quantized and executed on dedicated or domain-specific hardware accelerators.

The core computation behind DNNs are matrix multiplications (GEMMs) and convolutions, both of which are fundamentally composed by MAC operations, as presented in Figure~\ref{fig:hardware_diagram}(a). Consequently, enhancing the efficiency of multiplication and accumulation is crucial. In this context, shift-add-based multipliers ~\cite{judd2016stripes, hsu2020essa, rios2023bit, yu2024energy} are widely employed in hardware accelerators, as they can reduce area consumption, lower power costs, and relax critical timing constraints. Figure~\ref{fig:hardware_diagram}(b) illustrates an example of an 8-bit $\times$ 4-bit multiplication, yielding an 8-bit result via iterative shift-add operations. The input operands are in Q1.X format, that 1-bit for integer (i.e., 0 or -1) and others for fraction, ranging in (-1,1). The process begins at the least significant bit (LSB) of the multiplier, with each bit processed sequentially through an addition followed by a right-shift, and concludes at the most significant bit (MSB). The multiplication result is truncated to 8 bits, that is the same size of the multiplicand, since truncation loss is often minimal (only -0.12\% in this example).

However, while the shift--and-add scheme offers significant area and energy efficiency, it is constrained by increased latency, as a naive implementation requires n-cycle iterations for an n-bit multiplier operand. Consequently, reducing the size of the multiplier operand can substantially improve latency performance and further enhance energy efficiency. Although techniques such as executing multiple shift operations for trailing zeros (e.g., “1000”) within a single cycle~\cite{rios2019associativity,yu2024energy} or employing Canonical Signed Digit (CSD, e.g., 0111 to 100-) coding~\cite{yu2024energy} for the multiplier have been proposed to mitigate this issue, the overall latency remains strongly correlated with the size of the multiplier operand.  For DNNs, the multiplier operand typically corresponds to the weight. Therefore, using lower-bitwidth weights can eliminate latency concerns while simultaneously yielding additional reductions in energy consumption.

Therefore, in this work, we target a generic shift-add based arithmetic scheme to assess the benefits of our proposed quantization method relative to uniform quantization baselines, with the evaluation reflecting arithmetic efficiency rather than being bound to any particular hardware platform. More detailed discussion about model size, latency, energy consumption are provided in Section-VI.

\section{Methodology}
\label{sec:methodology}

\subsection{Algorithm Overview}
\label{subsec:algo_overview}
Based on our preliminary experiments and observations, \emph{SigmaQuant} proposes a two-phase procedure to determine per-layer quantization bitwidths, focusing first on a coarse clustering assignment and then on fine-grained adjustments. The goal of this section is to explain the structure and logic of this algorithm (summarized in Algorithm~\ref{alg:two_phase_quant}) without delving into low-level implementation or mathematical derivations. We take advantage of two key metrics – \textbf{\textit{weight standard deviation (Sigma)}} and \textbf{\textit{Kullback–Leibler (KL)}} divergence – to guide decisions (see Section~\ref{subsec:distribution_fitting} for the mathematical foundation and  preliminary experiments results of these metrics). In essence, Phase 1 provides an initial clustering-based bitwidth assignment, and Phase 2 provides an iterative, divergence-based refinement. Throughout the procedure, the algorithm tracks the accuracy and size of the model relative to these targets (with some buffer tolerance, $\Delta A$ for accuracy and $\Delta M$ for size) and uses two key metrics to guide decisions: weight standard deviation (denoted $\sigma$, a measure of the spread of the distribution of a layer) and Kullback-Leibler (KL) divergence between the original and quantized weight distributions. Algorithm~\ref{alg:two_phase_quant} outlines the overall procedure at a high level, summarizing the two phases described below.

\begin{enumerate} 
    \item \textbf{Phase 1 - Initial Phase (Cluster-Based Bitwidth Assignment):} The algorithm clusters the layers based on standard deviation $\sigma$ using an adaptive $k$-means method (with penalty parameter $\lambda$). Each cluster is mapped to a target bitwidth (e.g., from the set $\{2,4,6,8\}$). After quantization and calibration, the model’s accuracy and size are compared against the desired targets (with buffers $\Delta A$ and $\Delta M$). If neither metric is acceptable, the clustering is refined (by increasing $\lambda$) and the process repeats until at least one metric meets its boundary. 
    \item \textbf{Phase 2 - Refinement Phase (Iterative Improvement):} The algorithm fine-tunes the bitwidths of individual layers. A sensitivity score—combining $\sigma$ and the KL divergence between the full-precision and quantized weight distributions—is computed for each layer. Layers with high sensitivity (i.e., those most critical to accuracy) are adjusted by increasing their bitwidth, while layers with low sensitivity may have their bitwidth reduced to save memory. This iterative refinement continues until both accuracy and model size meet the targets, thereby placing the model in the Target Zone. 
\end{enumerate}

\begin{figure}[ht]
    \centering
    \includegraphics[width=1.0\linewidth]{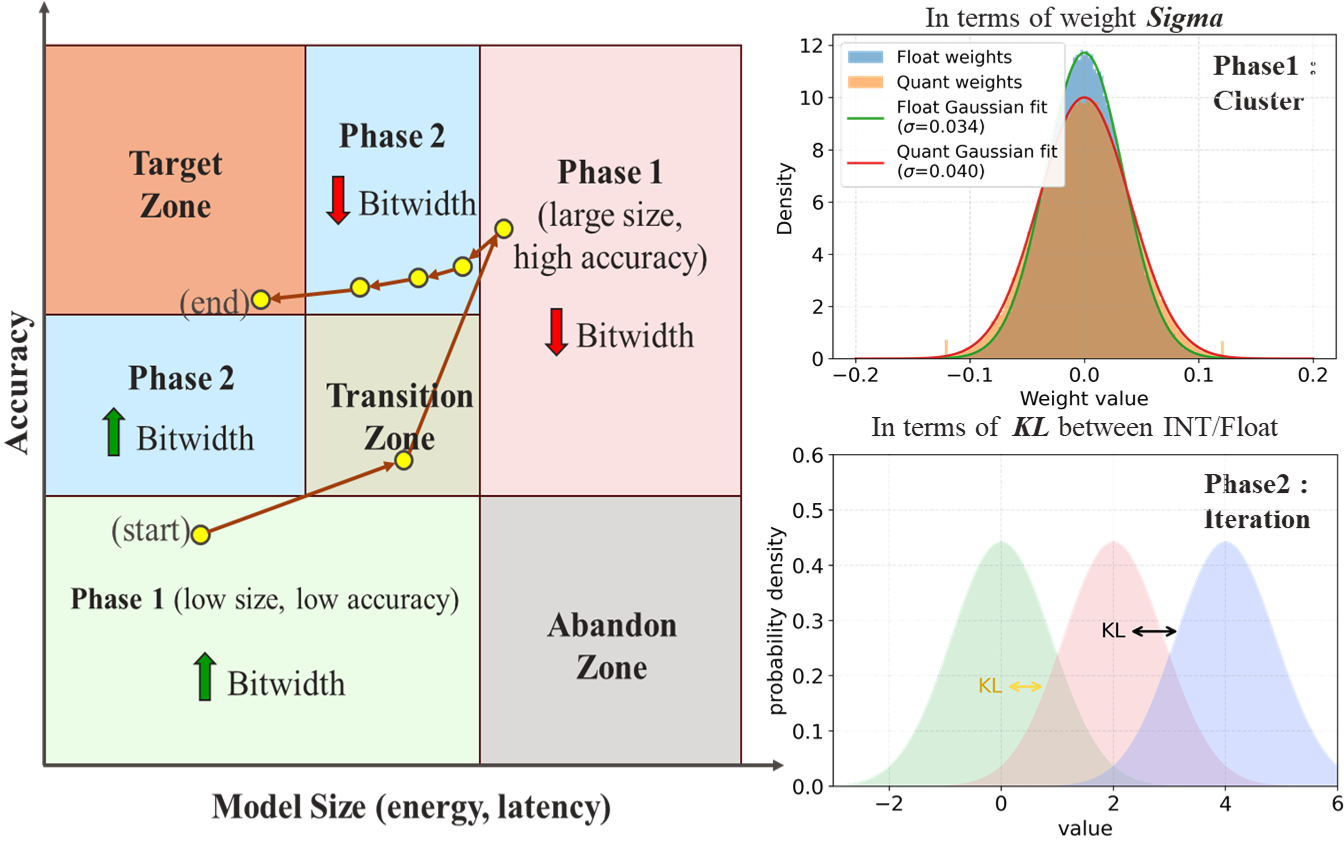}
    \caption{Overview of our proposed distribution-fitting quantization method. We start from user-defined boundary conditions (target model size and accuracy) and adapt bitwidths in two phases: initial clustering by standard deviation followed by iterative KL-based refinement.}
    \label{fig:method_diagram}
\end{figure}

Figure~\ref{fig:method_diagram} illustrates this two-phase procedure on a plot of model accuracy (y-axis) versus model size (x-axis). In the diagram, color-coded regions indicate the algorithm’s decision zones. We preserve those visual cues here to explain how the algorithm navigates the accuracy–model size trade-off:

\begin{itemize}[leftmargin=*,labelsep=0.4em,itemsep=0.2em,topsep=0.2em]
    \item \textbf{Bit Increase Zone (Phase 1 light green region):} 
    The model’s accuracy is too low, yet its current size is comfortably below the target budget. In this scenario, accuracy is the bottleneck. The algorithm responds by increasing bitwidths for certain layers (typically those most sensitive to quantization error) to regain accuracy

    \item \textbf{Bit Decrease Zone (Phase 1 light red region):} 
    The model’s accuracy is acceptable (at or above the required threshold), but the model size exceeds the target memory budget. Here, efficiency is the problem. The algorithm decreases bitwidths for some less-sensitive layers to compress the model and reduce its size

    \item \textbf{Transition Zone (middle gray-green region):} 
    Continue the previous cluster trend until reaching the next zone.
          
    \item \textbf{Abandon Zone (gray region):} 
    Neither accuracy nor model size is anywhere near acceptable levels. In this rare case, the procedure may terminate early to avoid wasting resources on a hopeless quantization scenario (i.e., the chosen constraints might be unattainable simultaneously). Terminate process.

    \item \textbf{Iteration Zone (Phase 2 region):} 
    This zone represents the intermediate state where one of the two criteria (accuracy or size) has met its target, but the other one has not yet met it. The model stays in this zone during the Phase 2 refinement process. The algorithm will perform iterative bitwidth tweaks – increasing or decreasing precision for selective layers – to improve the remaining metric while maintaining the already satisfied metric, effectively nudging the model toward the Target Zone. 

    \item \textbf{Target Zone (light orange region):} 
    Both accuracy and model size meet their target thresholds (within the allowed buffers). When the model lands in this zone, it means it satisfies the accuracy requirement and does not exceed the size budget.
\end{itemize}

\begin{algorithm}[!t]
\caption{\small Two-Phase Heterogeneous Quantization}
\label{alg:two_phase_quant}
\small          
\begin{algorithmic}[1]

\Require Float model $\mathcal{M}$ with $N$ layers; valid bit-set $\mathcal{B}=\{2,4,6,8\}$;
        targets $A_{\text{t}} , M_{\text{t}}$; tolerances $\Delta A , \Delta M$; max iters $I_{\max}$
\Ensure  Quantised model $\mathcal{M}_{q}$

\State $\mathcal{M}_{q} \gets \text{Initialize8Bit}(\mathcal{M})$   \Comment{start with uniform 8-bit}
\State $A \gets \text{Evaluate}(\mathcal{M}_{q})$
\State $M \gets \text{ModelSize}(\mathcal{M}_{q})$

\Statex\hspace{-\algorithmicindent}\textbf{--- Phase 1: adaptive clustering ---}

\State $\lambda \gets 0.1$;\; $i \gets 0$
\While{$(A < A_{\text{t}}-\Delta A) \;\land\; (M > M_{\text{t}}+\Delta M) \;\land\; (i<I_{\max})$}
    \State $i \gets i+1$
    \State $\textit{feat} \gets \text{StdDevFeatures}(\mathcal{M}_{q})$
    \State $\mathcal{C} \gets \text{AdaptiveKMeans}(\textit{feat},4,\lambda)$
    \State \text{AssignBitwidths}$(\mathcal{C},\mathcal{M}_{q},\mathcal{B})$
    \State \text{Calibrate}$(\mathcal{M}_{q})$;\; \text{QAT}$(\mathcal{M}_{q})$
    \State $A \gets \text{Evaluate}(\mathcal{M}_{q})$;\; $M \gets \text{ModelSize}(\mathcal{M}_{q})$
    \If{$(A \ge A_{\text{t}}-\Delta A) \lor (M \le M_{\text{t}}+\Delta M)$}
        \State \textbf{break} \Comment{one metric is inside the buffer}
    \Else
        \State $\lambda \gets \lambda + 0.1$
    \EndIf
\EndWhile

\If{$(A < A_{\text{t}}-\Delta A) \land (M > M_{\text{t}}+\Delta M)$}
    \State \Return{$\mathcal{M}_{q}$} \Comment{give up – infeasible}
\EndIf

\Statex\hspace{-\algorithmicindent}\textbf{--- Phase 2: iterative refinement ---}

\State $j \gets 0$
\While{$j < I_{\max}$}
    \State $j \gets j+1$
    \State \text{ImproveBitwidths}$(\mathcal{M}_{q})$
    \State \text{Calibrate}$(\mathcal{M}_{q})$;\; \text{QAT}$(\mathcal{M}_{q})$
    \State $A \gets \text{Evaluate}(\mathcal{M}_{q})$;\; $M \gets \text{ModelSize}(\mathcal{M}_{q})$
    \If{$(A \ge A_{\text{t}}) \land (M \le M_{\text{t}})$}
        \State \textbf{break}
    \EndIf
\EndWhile
\State \Return{$\mathcal{M}_{q}$}

\end{algorithmic}
\end{algorithm}


\subsection{Phase 1: Adaptive Clustering}
\label{subsec:phase1}

In the first phase, we assign initial bitwidths to each layer by clustering their standard deviations into $K=4$ groups, targeting for the bitwidths of 2-, 4-, 6-, 8-bit. For the initia assignment, we use the conventional k-mean. Then we check the location of the initial point in the Fig.~\ref{fig:method_diagram}, before we decide to increase the bitwidth or decrease the bitwidth. Next, unlike standard $k$-means, we introduce a adaptive term that discourages certain cluster from becoming too large, thus promoting a more uniform distribution of layers across the available bitwidths. Concretely, let $X = \{ x_1, x_2, \ldots, x_N \}$ be the set of $N$ layers we wish to cluster, where $x_i$ is a one-dimensional feature (e.g. standard deviation $\sigma_i$). We seek a partition
\[
  \mathcal{C} = \{C_1, C_2, \ldots, C_K\},
\]
of $X$ into $K$ clusters. We define the centroid of $C_j$ as
\[
  \mu_j \;=\; \frac{1}{|C_j|}\sum_{x \in C_j} x.
\]
The \emph{adaptive $k$-means} objective reads
\begin{equation}
\label{eq:balanced_kmeans}
  \min_{\substack{\mathcal{C},\,\mu_j}} 
  \sum_{j=1}^{K}
  \biggl[
    \sum_{x \in C_j} \|x - \mu_j\|^2 \;+\; \lambda 
      \Bigl(\,|C_j|\;-\;\tfrac{N}{K}\Bigr)^2
  \biggr],
\end{equation}
where $\lambda$ controls how strongly we penalize deviations from the ``ideal'' cluster size $\tfrac{N}{K}$. For each iteration:
\begin{itemize}[leftmargin=*,labelsep=0.4em,itemsep=0.2em,topsep=0.2em]
    \item We compute the distances from each layer $x_i$ to the centroid of each cluster, adjusted by the cluster-size penalty term $\lambda$. 
    \item We reassign $x_i$ to the cluster that minimizes its total cost.
    \item We update the centroids $\mu_j$ after all points are reassigned.
\end{itemize}
We initialize $\lambda$ to a small value (e.g., $0.1$) and gradually increase it (in increments, such as $0.1$) whenever the resulting bitwidth assignment fails to meet either of the global buffer conditions ($\Delta A$ and $\Delta M$ in \ref{alg:two_phase_quant}).
Each time a new clustering $\mathcal{C}$ is obtained, we map each cluster $C_j$ to a target bitwidth $b_j \in \{2,4,6,8\}$. We then calibrate and perform a short quantization-aware training (QAT) cycle. If neither accuracy nor model size usage falls within the acceptable range, we increment $\lambda$ and repeat. If, after a fixed maximum number of increments, both metrics remain unacceptably far from their targets, the algorithm halts (``abandoning'' the attempt and announcing the failure of quantization). We perform a quick calibration of the quantized model before QAT by using a subset of training data to adjust batchnorm statistics and quantization scales (ensuring stable QAT initialization).

\subsection{Phase 2: Iterative KL-Based Refinement}
\label{subsec:phase2}

Once Phase~1 ensures that at least \emph{one} of the target constraints (accuracy or model sizes) is satisfied, we proceed to a finer-grained iterative improvement of the per-layer bitwidths. Here, we focus on adjusting bitwidths to fix whichever metric remains suboptimal. We define a sensitivity measure for each layer $\ell$ that combines its standard deviation and a normalized Kullback--Leibler (KL) divergence between the float and quantized weight distributions.
We use the KL divergence as described in Equation~\ref{eq:kl}. We use a normalized version $\widehat{D}_{\mathrm{KL}}$ to bound it between 0 and 1, e.g. by dividing by $D_{\mathrm{KL}}\bigl(p_\ell \,\|\, p_{\text{int8}}\bigr)$ for an 8-bit baseline distribution $p_{\text{int8}}$.
In each iteration of Phase~2 we perform the following actions:
\begin{enumerate}
    \item \textbf{Measure Sensitivity:} For layer~$\ell$, we define a sensitivity score
    \[
      \widehat{D}_{\mathrm{KL}}(p_\ell\|\tilde{p}_\ell),
    \]
    Layers with a higher $\widehat{D}_{\mathrm{KL}}$ are more critical to accuracy, so if accuracy is below the target, we \emph{increase} their bitwidth first. Conversely, if model size usage is too high, we \emph{decrease} bitwidth on layers with a lower sensitivity score.
    \item \textbf{Apply Changes \& Calibrate:} We pick a small number of layers (e.g., 2 or 3) to adjust by $\pm2$ bits (within the allowable range $\{2,4,6,8\}$). We then recalibrate and run a short QAT cycle.
    \item \textbf{Re-Evaluate Metrics:} We check the global accuracy and model size usage. If both are within their buffers, we stop; if neither is acceptable, we keep adjusting.
    \item \textbf{Early Stopping / Reversion:} If too many consecutive adjustments fail to move the metrics closer to their goals, we revert to the previous stable assignment and exit.
\end{enumerate}
We use symmetric min--max range quantization for weights, performed per output channel (as implemented in Brevitas~\cite{brevitas}), which is a common hardware-friendly scheme. For activations, we adopt asymmetric quantization with statistical clipping at the 99.9th percentile to reduce outlier sensitivity and improve calibration robustness. Unless stated otherwise, we follow a memory-centric objective (Model Size counted on weights only), and thus keep activations fixed at 8 bits while allowing layer-wise weight bitwidths to vary within $\{2,4,6,8\}$ under SigmaQuant; when targeting compute (BOPs), both weights and activations can be adapted with the same procedure. Compared to recent MPQ methods that (i) co-design precision assignment with platform-specific inference units and pipelines (e.g., Edge-MPQ~\cite{Zhao2024EdgeMPQ}), (ii) rely on second-order sensitivity (Hessian) and specialized transition-aware QAT (e.g., HMQAT~\cite{Huang2025HMQAT}), (iii) explore a joint pruning--quantization space via reinforcement learning with an accelerator-level energy model~\cite{Balaskas2024DiversePruning}, or (iv) formulate bitwidth assignment as a global integer quadratic program with cross-layer dependency modeling (e.g., CLADO~\cite{Deng2025CLADO}), SigmaQuant adopts a lightweight distribution-fitting strategy. More specifically, Phase~1 clusters layers using $\sigma$ for a stable near-feasible initialization, and Phase~2 uses KL divergence to make small local bitwidth updates that explicitly enforce user-defined accuracy and resource constraints, without platform-specific co-design, Hessian estimation, RL exploration, or IQP solvers. By iterating this local adjustment approach, Phase~2 ensures that the model converges smoothly to a configuration that respects both accuracy and model size constraints without introducing large distributional mismatches at once; in practice, this step often accounts for relatively few bitwidth changes, given that Phase~1 has already established a near-feasible starting point.

\section{Experimental Setup}
\label{sec:experiments}

To validate the effectiveness of \emph{SigmaQuant} we conduct experiments on ImageNet~\cite{5206848} and CIFAR-100 datasets~\cite{cifar100} using InceptionV3~\cite{7780677} and five standard ResNet variants~\cite{7780459}---ResNet18, ResNet34, ResNet50, ResNet101, and ResNet152. We evaluate the proposed approach against state-of-the-art \emph{heterogeneous} quantization approaches as well as conventional \emph{uniform} quantization methods. To compare the result with other state-of-art method, we run experiments using the ImageNet (See Sec.~\ref{subsec:resnet_sota_memory}). To show the trend and relation between the model size and accuracy, we run an experiment using CIFAR-100 to save GPU resources (see Sec.~\ref{subsec:modelsize_accuracy_cifar100}).
We acquire the full-precision pre-trained models from the pytorch official website\cite{pytorchvisionmodels} for ImageNet. For CIFAR-100, we train these models again ourselves on the target dataset before quantization using either:
\begin{itemize}[leftmargin=*,labelsep=0.4em,itemsep=0.2em,topsep=0.2em]
    \item \textbf{Uniform Quantization:} All layers share a fixed bitwidth (e.g., 2, 4 or 8 bits).
    \item \textbf{SigmaQuant (Ours):} Layers receive bitwidths guided by their weight distribution properties (standard deviation and optional divergence checks).
\end{itemize}
Before quantization, a short calibration phase (using a small subset of the training set) is performed. Then QAT is used to mitigate performance degradation.

The metrics used to assess the different techniques are the following:
\begin{enumerate}
    \item \textbf{Model Size (MB):} Sum of quantized weights across all layers.
    \item \textbf{Top-1 Accuracy (\%):} Evaluated on the CIFAR-100 test set.
    \item \textbf{Regression Analysis:} For combined sets of quantized models, we fit accuracy--model-size curves and plot error bands (standard deviation) to visualize overall trends.
\end{enumerate}

Moreover, to validate the hardware impact from \textit{SigmaQuant}, we map the Resnet family on a general shift-add-based MAC unit and evaluate the latency and power consumption information through post-synthesis simulation with TSMC 28nm library.
Our experiments were run on NVIDIA A100 and V100 GPUs, and the results were validated through more than 8'500 GPU hours.

\section{Results}
\label{sec:results}

In this section, we first illustrate how the quantization process evolves over time in our two-phase algorithm, highlighting the transition from cluster-based assignment to iterative refinement. Then, we compare the performance of our method against state-of-the-art heterogeneous quantization schemes. Next, we analyze the trade-off between model size and accuracy in CIFAR-100, providing information on resource-accuracy balance. In the following, we discuss the hyperparameter effect. Finally, we demonstrate the potential hardware advantages.

\subsection{Phase-Based Learning Process}
\label{subsec:phase_process}

To illustrate the dynamic process of our method, we track the quantization states ("start" to "end") of a representative model (e.g., ResNet34) as it progresses through the two-phase algorithm described in Section~\ref{sec:methodology}.

\begin{figure*}[ht]
    \centering
    \includegraphics[width=1.0\linewidth]{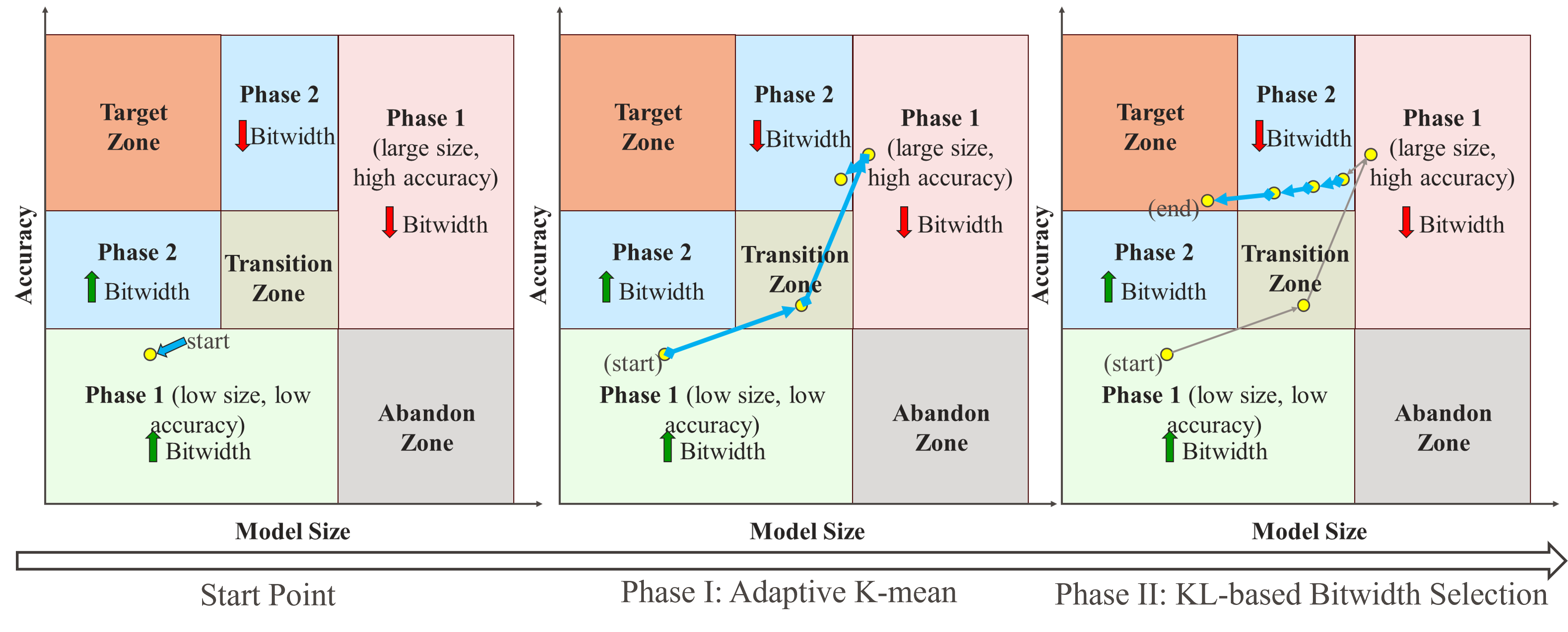}
    \vspace{-0.5cm}
    \caption{Example showing how training advances through the two-phase quantization for ResNet34. The x-axis represents corrected model sizes and the y-axis represents model accuracy. Different points indicate successive stages in the cluster phase (Phase 1) and iteration phase ((Phase 2), with the final quantized model landing in the target area.}
    \vspace{-0.5cm}
    \label{fig:resnet34_learning}
\end{figure*}

Figure~\ref{fig:resnet34_learning} plots the model accuracy (in \%) against the corrected model sizes (in megabyte), depicting the iterative exploration of the \emph{bit increase}, \emph{bit decrease}, and \emph{target} regions:
\begin{itemize}[leftmargin=*,labelsep=0.4em,itemsep=0.2em,topsep=0.2em]
    \item \textbf{Start Point – Conventional Clustering:}
    Initially, the bitwidth assignment is based on k-means clustering of layers. This coarse clustering provides an initial bitwidth assignment that quickly improves accuracy while causing the model size to approach the target buffer ( Fig.~\ref{fig:resnet34_learning} left).
    \item \textbf{Phase I – Iterative Re-Clustering:}
    In Phase I, the model re-clusters individual layer bitwidths using an adaptive clustering method (Eq.~\ref{eq:balanced_kmeans}). This technique assigns bitwidths to a higher or lower cluster, transitioning the model from the bit-increasing zone to the bit-decreasing zone and subsequently into Phase 2 (Fig.~\ref{fig:resnet34_learning} middle).
    \item \textbf{Convergence and Abort Conditions:}  
    Once one of the target metrics (accuracy or model size) meets the buffer threshold, Phase 2 commences. Here, we compute a sensitivity score for each layer that combines its weight standard deviation with a normalized Kullback–Leibler divergence measure. Layers exhibiting minimal divergence, indicating that further precision reduction would incur only minor accuracy loss, are selectively adjusted (Fig.~\ref{fig:resnet34_learning} right).
\end{itemize}

We fix the number of clusters to 4, corresponding to weight bitwidths of $\{2,4,6,8\}$-bit. These choices are illustrative, and other configurations could equally be adopted. Phase~1 runs up to 2 iterations (this number can be configured to a higher value when facing larger models), each followed by 4 epochs of QAT. Phase~2 allows up to 40 refinement steps with 40 QAT epochs per step, adjusting 2 layers (we fixed this to 2 layers in our implementation for all experiments) per iteration based on sensitivity. QAT uses cross-entropy loss and SGD (for ResNet) or Adam (for others), with a reduced learning rate. We start from INT8 quantized models. The accuracy threshold is set to 1\% drop, and the memory constraint targets 75\% of the INT8 model size ($\approx$18.75\% of the FP32 baseline). 
In general, Figure~\ref{fig:resnet34_learning} demonstrates the stability of the quantization path, showing that only a handful of iterations are typically required before a near-optimal trade-off between accuracy and model size overhead.
Additionally, we compare the configuration obtained after Phase I alone with the final configuration after Phases I–II, to quantify the added benefit of Phase II. Table~\ref{tab:phase_com} summarizes results across models under a $\leq$2\% accuracy‑drop constraint and a $\leq$40\% INT8‑size budget. As noted, Phase 1—our fast, adaptive k-means clustering based solely on layer‑wise standard deviation—can leave the model slightly above or below the memory target; the table therefore reports the Phase 1 (“std‑only”) accuracy/size alongside the final result after Phase 2. After Phase 1 we choose a direction: either increase the bitwidth of the most sensitive layers to recover accuracy or decrease bitwidths where possible to meet the size budget. This operating region near 35–40\% of the INT8 size is particularly challenging because further bitwidth reductions leave limited headroom to preserve accuracy. For ResNet‑18, Phase 1 already satisfies the size budget but the accuracy drop is excessive; increasing the precision of selected layers (\umark in Table~\ref{tab:phase_com}) restores accuracy while keeping the model within budget, so both constraints are ultimately met. For ResNet‑34, both constraints are met after Phase 1, so no refinement is required. For ResNet‑50, the size target is met but the accuracy constraint remains violated even after Phase 2, indicating that the two targets cannot be satisfied simultaneously under this setting.
\begin{table}[t]
\centering
\caption{Model sizes and accuracies.}
\begingroup
\setlength{\tabcolsep}{4pt}
\resizebox{\linewidth}{!}{%
\begin{tabular}{@{}lcccccccc@{}}
\toprule
Model & \makecell{Int8\\Size\\(MiB)} & \makecell{Int8\\Acc.\\(\%)} &
\makecell{Final\\Acc.\\(\%)} & \makecell{Final\\Size\\(MiB)} &
\makecell{Phase\,I\\Acc.\\(\%)} & \makecell{Phase\,I\\Size\\(MiB)} &
\makecell{Next\\Phase} & \makecell{Target\\Meet} \\
\midrule
\meetrow
ResNet18  & 11.14 & 80.40 & 78.41 & 3.72 & 75.98 & 3.20 & \umark & \textcolor{green}{\cmark} \\
\meetrow
ResNet34  & 20.77 & 82.90 & - & - & 81.14 & 6.52 & - & \textcolor{green}{\cmark} \\
\failrow
ResNet50  & 24.32 & 84.01 & 81.21 & 8.49 & 82.57 & 8.93 & \dmark & \textcolor{red}{\xmark} \\
\meetrow
ResNet101 & 42.38 & 85.98 & 84.31 & 12.28 & 84.78 & 15.26 & \dmark & \textcolor{green}{\cmark} \\
\failrow
ResNet152 & 57.26 & 86.77 & 84.67 & 16.72 & 85.26 & 22.66 & \dmark & \textcolor{red}{\xmark} \\
\bottomrule
\end{tabular}%
}
\endgroup
\label{tab:phase_com}
\end{table}

\subsection{Performance Comparisons with State-of-the-Art Heterogeneous Quantization Schemes}
\label{subsec:resnet_sota_memory}

In this subsection, we compare our proposed quantization technique (\textbf{Ours}) against state-of-the-art methods on ResNet-50~\cite{resnet} and InceptionV3~\cite{7780677}, focusing on model size (in MB) and Top-1 accuracy (\%). We implement our method to achieve optimal performance under various memory constraints by setting target bitwidths to 2, 4, 6, and 8, resulting in a mixed-bitwidth configuration. Table~\ref{tab:resnet-sota} consolidates results from prior work~\cite{Baskin2021uniq,mishra2018apprentice,zhou2017incremental,xu2018deep,polino2018model,yao2021hawq, Deng2025CLADO} alongside our heterogeneous bitwidth assignments for weights (W) and activations (A) (column Bits(W,A) in the table). We emphasize weight precision while treating activations as 8-bit, since our primary objective is memory reduction. For each model, we list the baseline full-precision size and accuracy, then compare uniform or mixed-precision baselines to illustrate how different bit configurations trade off memory and accuracy. The methods labeled Ours highlight two representative configurations that demonstrate a favorable balance between model compactness and classification performance.

\begin{table}[ht]
\centering
\caption{Comparison of quantization methods on InceptionV3 and ResNet-50.}
\begin{tabular}{lccc}
\toprule
\textbf{Method} & \textbf{Bits(W,A)} & \textbf{Model Size(MB)} & \textbf{Top-1 Acc.(\%)} \\
\midrule
\multicolumn{4}{c}{\textbf{ResNet-50}} \\
\midrule
Baseline \cite{yao2021hawq}       & 32,32   & 97.8  & 77.72 \\
Apprentice \cite{mishra2018apprentice}   & 2+, 8+  & 14.11   & 72.8  \\
UNIQ \cite{Baskin2021uniq}          & 4,8     & 12.8    & 74.37 \\
UNIQ \cite{Baskin2021uniq}          & 4+, 8+  & 12.8    & 75.1  \\
Apprentice \cite{mishra2018apprentice}   & 4+, 8+  & 20      & 74.7  \\
Apprentice \cite{mishra2018apprentice}   & 2+,32   & 14.1    & 74.7  \\
UNIQ \cite{Baskin2021uniq}          & 4,32    & 12.8    & 75.09 \\
HAWQ-V3 \cite{yao2021hawq}     & 8,8     & 24.5    & 77.58 \\
HAWQ-V3 \cite{yao2021hawq}     & 4/8,4/8 & 18.7    & 76.73 \\
HAWQ-V3 \cite{yao2021hawq}     & 4,4     & 13.1    & 74.24 \\
CLADO \cite{Deng2025CLADO}     & mix, 8   & 17.89   & 75.42 \\
CLADO \cite{Deng2025CLADO}     & mix, 8   & 13.42   & 73.10 \\

\rowcolor{yellow!20}
Ours                      & mix,8   & 12.02   & 76.86 \\
\rowcolor{yellow!20}
Ours                      & mix,8   & 10.78   & 75.63 \\
\midrule[1.5pt]
\multicolumn{4}{c}{\textbf{InceptionV3}} \\
\midrule
Baseline \cite{yao2021hawq}         & 32,32   & 90.9  & 78.88 \\
Integer Only \cite{Jacob2018Quantization}   & 8, 8  & 22.7   & 74.20  \\
Integer Only \cite{Jacob2018Quantization}   & 7, 7  & 20.1   & 73.70  \\
RVQuant \cite{Baskin2021uniq}          & 8,8     & 22.7   & 74.22  \\
HAWQ-V3 \cite{yao2021hawq}     & 4/8,4/8 & 19.6    & 74.65 \\
\rowcolor{yellow!20}
Ours                      & mix,8   & 19.63   & 74.73 \\
\bottomrule
\end{tabular}
\label{tab:resnet-sota}
\end{table}

Focusing primarily on memory efficiency, these results show that our heterogeneous quantization yields models that are up to \textbf{7--10$\times$ smaller} than the full-precision baseline, while retaining most of the original accuracy. For example, for ResNet-50, \emph{Ours (mix,8)} (12.02~MB) exceeds 76\% accuracy, aligning with methods that use similar or slightly larger model sizes, while 13.1MB for HAWQ-V3 with 74.24\% accuracy, 13.42MB for CLADO with 73.10\% accuracy. Similarly, for InceptionV3, our method reach similar Model Size (19.6MB for HAWQ-V3, 19.63MB for Ours) and Top-1 accuracy (74.65\% for HAWQ-V3, 74.73\% for Ours) as the HAWQ-V3. Notably, we achieve these gains using a straightforward procedure that incorporates standard deviation and divergence metrics to adapt the bitwidth per layer. Compare to the HAWQ-V3, our method support more bitwidths choices, and it can shrink the model size adaptively according to the memory constrain (as shown in Fig.\ref{fig:modelcompare}(b)). These comparisons confirm that our method effectively balances memory footprint with competitive accuracy, making it a suitable choice for scenarios where model size is the foremost constraint.
In our design, SigmaQuant’s search cost is dominated by short QAT loops rather than by an expensive discrete search. Phase~1 computes per-layer statistics (one scalar $\sigma$ per layer) and runs adaptive $k$-means on these $L$ layers with $K=4$ clusters, which is limited in practice (M iterations for example). Computing KL divergences uses layer histograms and scales with the number of parameters $P$ once per refinement round. Phase~2 makes small local moves---changing $m$ layers per round ($m=2$ in our setup)---and applies brief QAT to re-stabilize accuracy. Overall, the wall-clock is therefore well approximated by the number of QAT epochs:
\[
\mathrm{Cost} \approx \bigl(M\,E_{\text{P1}} + N\,E_{\text{P2}}\bigr)\times T_{\text{epoch}},
\]
where $M$ is the number of phase~1 rounds, $E_{\text{P1}}$ are Phase~1 epochs, $N$ is the number of refinement rounds, $E_{\text{P2}}$ are epochs per round, and $T_{\text{epoch}}$ is the time for one QAT epoch. In our experiments we cap Phase~1 at normally 1 to 3 iterations and Phase~2 runs normally within 5 (smaller models) to 40 (large models) refinement rounds, with early stopping once both accuracy and size targets are met. As a result, end-to-end times are model-dependent: for the runs in this paper we measured $\approx$2 to 30 hours for ResNet-18/34/50/101/152, respectively, using A100/V100 GPUs (see Sect.~\ref{subsec:modelsize_accuracy_cifar100}). We do not claim to be faster than calibration-only PTQ methods that avoid any fine-tuning; instead, SigmaQuant trades a moderate offline search for two properties that those methods typically lack: (i) hard-constraint adaptivity (meeting user-specified accuracy/size targets across devices) and (ii) consistently better accuracy--size and hardware PPA trade-offs (Tables~\ref{tab:phase_com}~\ref{tab:resnet-sota}, Figs.~\ref{fig:modelcompare}).
Methods such as HAWQ-V3 estimate per-layer sensitivity via second-order information and solve a global assignment with ILP; their runtime is dominated by many backward passes for Hessian/spectrum estimation plus calibration, after which assignment is relatively cheap. UNIQ and Apprentice require training (noise injection or distillation) across multiple epochs; they do not perform a per-layer mixed-precision search but still incur substantial wall-clock due to full-model optimization.

SigmaQuant lies in between: it avoids Hessian estimation and global RL/ILP search, yet it still adapts layer precisions to explicit hardware/accuracy constraints via a small number of targeted QAT loops. In short, SigmaQuant is not a ``zero-search'' PTQ method; it is an adaptive mixed-precision method with linear-in-layers bookkeeping and a bounded number of short QAT epochs, which empirically yields favorable accuracy--size--hardware trade-offs under tight device constraints.

\subsection{Model Size and Accuracy Analysis}
\label{subsec:modelsize_accuracy_cifar100}

\begin{figure*}[ht]
    \centering
    \includegraphics[width=1.0\linewidth]{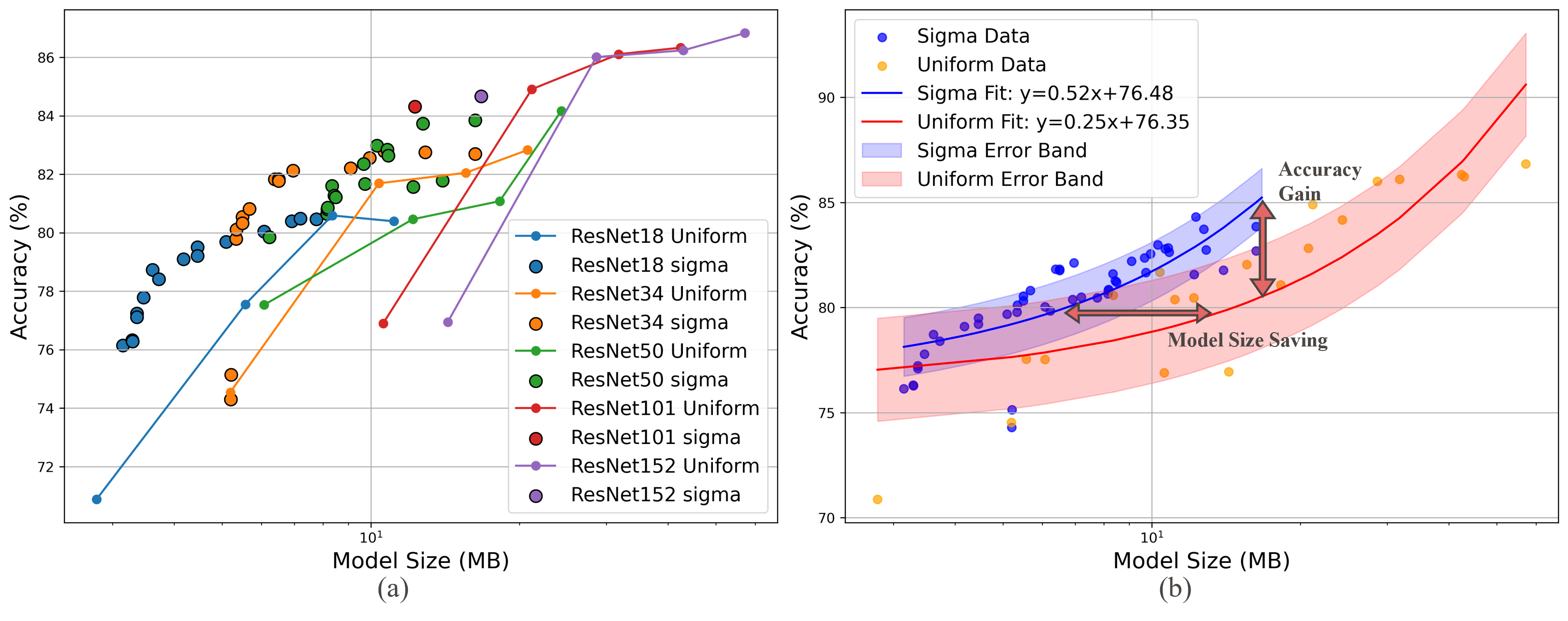}
    \vspace{-0.5cm}
    \caption{(a) Comparison of Top-1 accuracy versus model size for various ResNet architectures on CIFAR-100, where darker markers denote the \emph{sigma}-based method and lighter markers denote uniform quantization. (b) Regression fits with ±1\ensuremath{\sigma} error bands reveal that the \emph{sigma} approach consistently achieves higher accuracy at equivalent model sizes.}
    \vspace{-0.5cm}
    \label{fig:modelcompare}
\end{figure*}

Another key advantage of our method is its adaptive nature. By dynamically adjusting the boundary conditions for both the model size buffer and the accuracy target, our approach performs layer-wise quantization in a highly flexible manner, a capability that previous methods lack. Figure~\ref{fig:modelcompare} (a) compares model size and Top-1 accuracy on CIFAR-100 for five different ResNet architectures (ResNet-18, ResNet-34, ResNet-50, ResNet-101, and ResNet-152) under two quantization schemes: conventional uniform quantization and our \emph{SigmaQuant} approach. Across all model variants, \emph{SigmaQuant} (plotted as darker markers) achieves higher accuracy for a given model size than their uniformly quantized counterparts (lighter markers). In practical terms, for an equal memory footprint, a model quantized with \textit{SigmaQuant} yields 4\% more accuracy. For the same accuracy, the model size can shrink by around 40\%. Notably, the uniform quantization approach is unable to reach the optimal balance between model size and accuracy, highlighting its inherent limitations. This trend holds consistent as networks become deeper: model quantized with our method surpasses the accuracy of a similarly-sized uniformly quantized one. For example, the ResNet50 sigma can achieve similar accuracy (81.0\% for uniform quantization, 81.7\% for sigma quantization) with better model size (18.24MB for uniform quantization, 13.98MB for sigma quantization). In effect, \textit{SigmaQuant} retains and, in some cases even improves, the predictive performance while aggressively reducing the size of the model.

To further illustrate the overall trend, Fig.~\ref{fig:modelcompare} (b) consolidates the results by plotting top-1 accuracy against model size for all quantized ResNet models tested, with linear regression fits for each quantization scheme (sigma-based vs. uniform) and shaded ±1\ensuremath{\sigma} error bands. The key observation is that the fitted curve for \textit{SigmaQuant} lies consistently above that of uniform quantization across the entire range. The separation between these trendlines is substantial, and the error bands show minimal overlap, indicating that the accuracy gap is statistically robust across multiple runs. In other words, \textit{SigmaQuant} achieves a given accuracy level with fewer parameters (3.2MB \textit{Model Size Saving} in Fig.~\ref{fig:modelcompare} (b)), or conversely, for the same number of parameters (same memory budget), it delivers higher accuracy than uniform precision (4\% \textit{Accuracy Gain} in Fig.~\ref{fig:modelcompare} (b)). This translates to a more favorable accuracy–efficiency trade-off. Notably, the advantage becomes even more pronounced for larger models: \textit{SigmaQuant}’s accuracy approaches that of higher-precision (e.g., 32-bit) networks at a fraction of the memory usage as model size grows. Such behavior is highly desirable for resource-constrained deployments, as it means that one can enjoy near-baseline accuracy without the cost of a large model. Overall, the regression fits and non-overlapping error bands indicate that SigmaQuant consistently provides a better accuracy–model-size trade-off than uniform quantization across different ResNet depths.

Our experiments also demonstrate that SigmaQuant remains practical for offline training across the ResNet family. For example, under an extreme constraint, limiting the model size to achieve at least 25\% memory savings while tolerating at most 1\% accuracy drop relative to the full precision model, the total quantization search and training times were approximately 4, 2, 1.5, 5.5, and 26 hours for ResNet-18, ResNet-34, ResNet-50, ResNet-101, and ResNet-152, respectively. These runtimes remain within a reasonable range for offline deployment. Even in cases where the training process fails due to overly tight memory constraints, we observe from the generated models that SigmaQuant still produces meaningful accuracy–efficiency trade-offs, as shown in the corresponding evaluation figures.

\subsection{Hyperparameter and the Activation}
\label{subsec:Hyper_Act}
We study the impact of the accuracy/size buffers $(\Delta_A,\Delta_M)$ and the refinement schedule (max rounds $N_{\max}$, step size $m$ layers/round, and QAT epochs/round $E_{\text{P2}}$) on convergence and wall‑clock.
Recall that Phase~1 ends once \emph{either of} the metrics enters its buffer (Alg.~1, lines 12–16), whereas Phase~2 continues until both strict targets are met (lines 27–31). Thus, larger buffers generally reduce Phase~1 iterations but may require more Phase~2 refinements; smaller buffers have the opposite effect. We keep $K{=}4$ clusters and the default 2‑layer step in Phase~2 unless otherwise stated.
Table~\ref{tab:sensitivity} summarizes typical behavior on \textbf{ResNet‑34/CIFAR‑100} under our default targets (at most 1\% accuracy drop), varying only $(\Delta_M)$ and refinement settings. The trends hold qualitatively for other models.
Overall, smaller buffers (\emph{Conservative}) reduce the chance of overshooting a target but increase observed rounds $N$ and wall‑clock; larger buffers (\emph{Aggressive}) do the opposite, at a slight risk of needing extra micro‑adjustments later in Phase~2. In all cases, the final solution respects the strict targets due to the Phase~2 stopping rule.

\begin{table}[h]
\centering
\caption{Sensitivity of SigmaQuant on ResNet-34 (CIFAR-100) under the default targets.}
\label{tab:sensitivity}
\resizebox{\linewidth}{!}{%
\begin{tabular}{@{}lcccccc@{}}
\toprule
Setting & $\Delta_A$ & $\Delta_M$ & Obs.\ $M$ & Obs.\ $N$ & Time (h) & Meet? \\
\midrule
Conservative       & 1\% & 85\% & 3 & 0 & $\sim$4.5  & \cmark \\
Balanced (default) & 1\% & 75\% & 3 & 3 & $\sim$12.6 & \cmark \\
Aggressive         & 1\% & 50\% & 4 & 5 & $\sim$19.0 & \xmark \\
\bottomrule
\end{tabular}%
}
\end{table}

Moreover, except the weights for the memory saving, we adapt the activation to reduce the amount of BOPs. To evaluate activation reduction fairly, we switch the target from memory to compute:
\[
\textstyle \mathrm{BOPs} \triangleq \sum_{\ell} B_w(\ell)\,B_a(\ell)\,\mathrm{MACs}(\ell),
\]
where the $B_w(\ell)$ and $B_a(\ell)$ are the bitwidth of weight and activation; $\mathrm{MACs}$ is the multiply–accumulate operation of this layer. It upper‑bounds bit‑level work and correlates with energy/latency (on our shift‑and‑add MACs, cycles scale primarily with $B_w$) while exposing the benefit of lowering $B_a$. 
We then run SigmaQuant with a compute budget (BOPs) and an accuracy target, letting both weights and activations adapt.
Table~\ref{tab:act} reports indicative results on \textbf{AlexNet/ResNet‑18/ResNet‑32/ResNet‑50} on CIFAR‑100 with a 1\% accuracy‑drop target and a 25–35\% BOPs‑reduction budget. As expected, activation down‑quantization reduces BOPs but leaves Model Size (MB) unchanged; SigmaQuant co‑adapts layerwise weights to maintain accuracy.
When the optimization target is memory (weight size), changing activation bitwidth has no effect on Model Size (MB) because activations are not counted in that metric. Under a compute target (BOPs), SigmaQuant naturally co-optimizes toward lower activation precision while preserving accuracy, yielding 35–50\% BOPs reductions in our indicative runs with $\leq$2.5\% accuracy loss.

\begin{table}[h]
\centering
\caption{Activation reduction under a BOP target.}
\label{tab:act}
\begin{tabular}{lcccc}
\toprule
Model & $\Delta_{A}$ & $\Delta_{\text{BOP}}$ \\
\midrule
ResNet‑18 & 78.91\% & ($-32.9\%$) \\
ResNet‑34 & 81.31\% & ($-49.4\%$) \\
ResNet‑50 & 82.38\% & ($-32.0\%$) \\
\bottomrule
\end{tabular}
\end{table}


Overall, these results show that our distribution-guided \textit{SigmaQuant} method provides an effective balance between model compression and predictive performance. By judiciously allocating bitwidths based on per-layer weight statistics and distributional divergence, \textit{SigmaQuant} consistently produces more compact models without sacrificing accuracy under the same model-size constraints. This represents a compelling improvement over uniform quantization, making our approach well-suited for real-world DNN deployment in memory-limited embedded environments.

\subsection{Hardware Performance Analysis}
\label{sec:hw_impl}

\vspace{-0.15cm}
\begin{table}
  \centering
  \caption{MAC Implementations}
  \label{Tab:MACHW}
  \begingroup
  \renewcommand{\arraystretch}{1} 
  \resizebox{1\linewidth}{!}{%
    \begin{tabular}{@{}lccccc@{}}
      \toprule
      & \textbf{FP32} & \textbf{FP16} & \textbf{BF16} & \textbf{INT8} & \textbf{Shift-add} \\
      \midrule
      Multiplication &
        \makecell{32-bit\\1 subword\\FP32$\times$} &
        \makecell{32-bit\\2 subwords\\FP16$\times$} &
        \makecell{32-bit\\2 subwords\\BF16$\times$} &
        \makecell{32-bit\\4 subwords\\INT8$\times$} &
        \makecell{32-bit\\4 subwords\\8-bit $\gg$+} \\[12pt]
      Accumulation &
        \makecell{1 subword\\FP32$+$} &
        \makecell{2 subwords\\FP32$+$} &
        \makecell{2 subwords\\FP32$+$} &
        \makecell{4 subwords\\INT32$+$} &
        \makecell{4 subwords\\INT32$+$} \\[12pt]
      Area / $\mathrm{\mu m^{2}}$ &
        \makecell{3218.3} &
        \makecell{3837.9} &
        \makecell{3501.9} &
        \makecell{2103.4} &
        \makecell{1635.4} \\
      \bottomrule
    \end{tabular}%
  }
  \endgroup  
  \label{tab:arith-compare}
\end{table}
\vspace{-0.15cm}

\begin{figure*}[ht]
    \centering
    \includegraphics[width=1.0\linewidth]{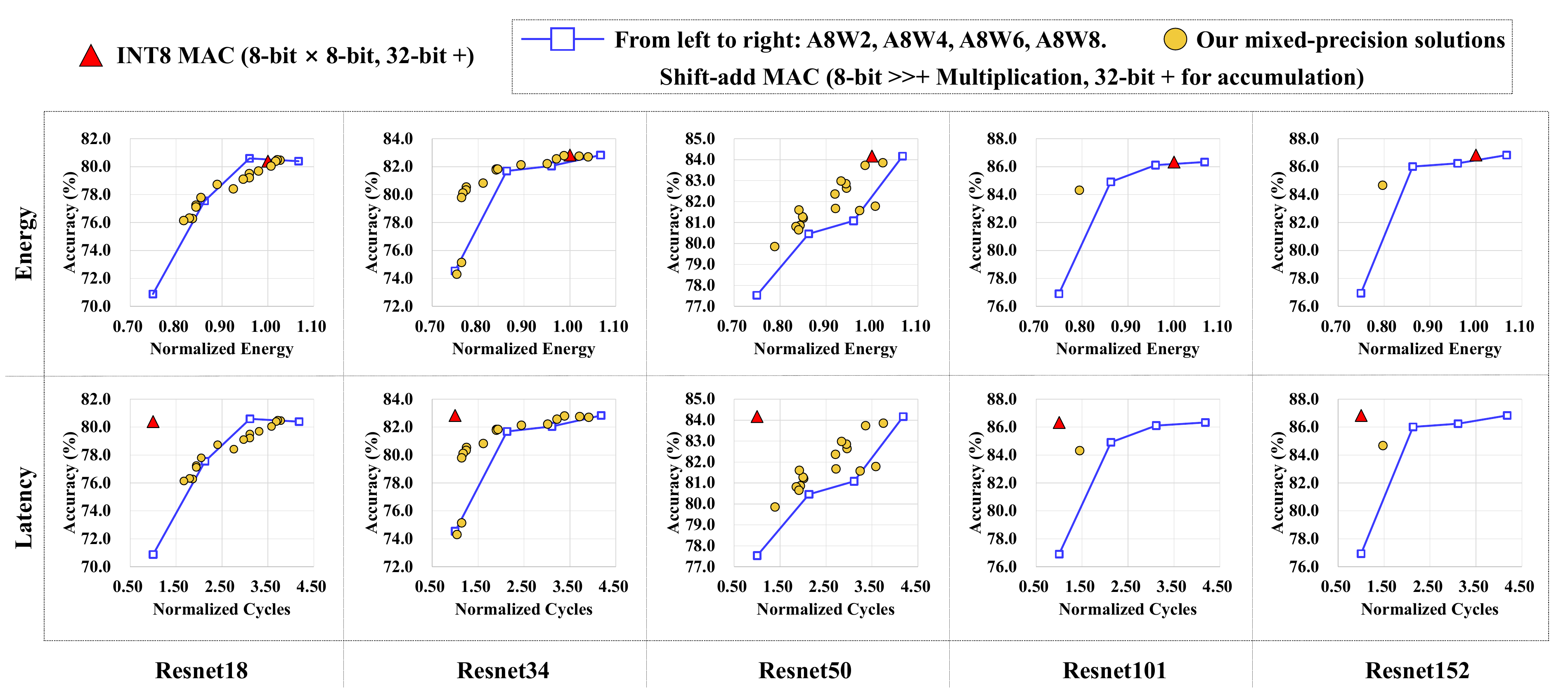}
    \caption{ Normalized energy consumption (top) and cycle count (hence latency, bottom)} versus accuracy for all ResNet models with uniform quantization(A8W2, A8W4, A8W6, A8W8) and sigma quantization schemes. An INT8 MAC, which has 1-cycle 8-bit $\times$ 8-bit multiplier and 32-bit adder, is used as a baseline for performance comparison. FP32, FP16, and BF16 alternatives are not presented here since they incurs huge overhead , that up to 5.5$\times$, 4.0$\times$, and 3.6$\times$ more energy cost. All the data are normalized with INT8 computation. The closer a dot element is to the top-left corner, the better its performance and accuracy.
    \label{fig:Energy_delay_accuracy}
\end{figure*}

A key advantage of heterogeneous quantization is that it leverages the varying robustness of different layers to maintain accuracy while allowing the corresponding hardware to operate with less area occupation, fewer cycle count, and lower energy consumption, increasing all benefits.


In this work, we target a widely used shift-add MAC implementation to evaluate the performance under different quantization schemes. This MAC unit has an 8-bit shift-add-based multiplier that supports $8\text{-bit} \times n\text{-bit} \rightarrow 8\text{-bit}$ multiplication through iterative right-shift and addition, followed by a 32-bit adder for accumulation. The shift-add design performs a single addition and multiple right-shift operations within each cycle, which enables to process trailing zeros of the multiplier operand, reducing the average latency to roughly n/2 cycles for an n-bit operand. The shift-add MAC hardware is implemented in TSMC 28nm technology (0.9V, 600MHz) and features a 32-bit datapath. It should be emphasized that the arithmetic performance comparison reflect a general case applicable to all technology nodes and hardware platforms, while the presented implementation is intended primarily for illustrative purposes. We also implemented FP32, FP16, BF16, and INT8 alternatives under the same conditions, detailed in Table~\ref{Tab:MACHW}. It can be found that shit-add MAC implementation reduces 22.3\% area over the INT8 one, and more than 49.2\% over others. We characterize the performance using post-synthesis simulation, mapping all convolutional and fully-connected layers of ResNet models onto the hardware. The energy consumption and cycle count for inference are evaluated, as shown in Figure~\ref{fig:Energy_delay_accuracy}.
In Figure~\ref{fig:Energy_delay_accuracy}, we show the normalized energy consumption (top, a) and cycle count (hence latency, (bottom, b)) of different quantization schemes across benchmarks. Both our \emph{SigmaQuant} scheme and the uniform quantization approach are mapped on the shift-add design, whereas the typical INT8 quantization is performed with the INT8 hardware that already has high efficiency. The results of FP32, FP16, BF16 are not included, since they incur up to 5.5$\times$, 4.0$\times$, and 3.6$\times$ more energy cost over the INT8 one, respectively. The uniform quantization approach has four combinations of 8-bit activations and 2/4/6/8-bit weights, namely as A8W8, A8W6, A8W4 and A8W2. All the values are normalized by the INT8 MAC implementation. The accuracy drop is defined as the difference between FP32 reference and the quantized ones.

As expected, in uniform quantization, when weights reduce from 8-bit to 4-bit, and especially to 2-bit, the accuracy dramatically drops. Taking Resnet34 as an example, even if uniform quantization A8W2 saves 25.0\% energy consumption wrt. the INT8 case, it degrades accuracy by 8.54\%. However, \emph{SigmaQuant} performs a larger search in the design space and is able to address this issue, achieving 23.3\% energy savings over the INT8 alternative with only a 2.97\% accuracy loss. As for the A8W4 case, uniform quantization can save 13.8\% energy consumption with 1.39\% accuracy loss while our method reduces 16.0\% energy cost with only 1.25\% accuracy degradation. Furthermore, the results of \emph{SigmaQuant} are consistent across all the benchmarks, always yielding data points located closer to the upper left corner in the figure wrt. the uniform approach, hence less latency and energy consumption with lower accuracy loss. The only exception is ResNet18, where \emph{SigmaQuant} shows a very similar trend to uniform quantization, since such model is too small and therefore the space search is limited. Still, \emph{SigmaQuant} can get 15.7 \% less energy consumption wrt. the INT8 case with 3.76\% accuracy drop while that of uniform quantized one is 13.8\% and 3.45\%. For larger models such as Resnet101 and Resenet152, that provides enough searching space, our \emph{SigmaQuant} method can reduce up to 20.6\% and 20.3\% energy cost wrt. the INT8 alternative, which is already a well-known solution, with comparable accuracy. Note that the model size of our method is always smaller than that of INT8 quantization for all the comparisons.

Similar trend can be observed in the latency performance. Due to the serial shift-add-based multiplication, even multiple shifts per cycle is used, it still incurs latency overhead. For example, the A8W8 uniform quantization of ResNet34, runs 4.2$\times$ slower than the INT8 case. Thanks to the fully exploitation and optimization by \emph{SigmaQuant}, however, it dramatically cuts this overhead. For Resnet34, it lowers energy by 23.3\% while increasing latency by only 17.5\% relative to the INT8 alternative. This analysis only considers the general shift-add arithmetic, yet the advantages would become far more pronounced with specific techniques such as CSD encoding~\cite{yu2024energy} applied to the multiplier operand.

In addition, uniform quantization offers a very limited range of choices, making it difficult to balance accuracy with latency and energy efficiency. In contrast, the \emph{SigmaQuant} dot distribution shown in Figure~\ref{fig:Energy_delay_accuracy} illustrates a broader set of quantized models. For example, the options between A8W2 to A8W4 or A8W4 to A8W6 for ResNet18, ResNet34, and ResNet50. This greater flexibility allows for more effective tailoring to diverse hardware requirements, including model size, runtime latency, and energy budget.

Overall, these results demonstrate that distribution-guided, layer-wise bit allocation can significantly enhance hardware efficiency compared to uniform quantization. In fact, our approach shows greater improvements than the A8W4 variant, approaching the efficiency of A8W2 quantization while incurring minimal accuracy loss, and can further improve hardware efficiency than typical INT8 solution. As model size decreases, fine-grained quantization leverages the shift-add architecture to achieve lower latency and reduced power consumption, offering a compelling solution for resource-constrained deployments.

\vspace{-0.15cm}
\section{Conclusion}
\label{sec:conclusion}
\vspace{-0.15cm}

In this work, we have presented \textbf{\textit{SigmaQuant}}, an adaptive, layerwise heterogeneous quantization framework that leverages a distribution-fitting approach to assign bitwidths based on both weight standard deviation and KL divergence. Our two-phase method, featuring an initial cluster-based bitwidth assignment followed by an iterative, divergence-driven refinement, effectively balances accuracy and model size with a reasonable searching effort, ensuring that quantized models maintain high performance while achieving significant compression.
A key strength of \textit{SigmaQuant} is its inherent adaptability. Across different platforms with varying constraints and requirements, our framework consistently adapts bit allocation to satisfy specific resource budgets. For example, by integrating with hardware-centric metrics such as model size and target accuracy, \textit{SigmaQuant} directly translates per-layer quantization decisions into efficient bitwidth assignments that align with the overall requirements of the system.
In general, our experimental results on CIFAR-100 and ImageNet across multiple DNN architectures demonstrate that \textit{SigmaQuant} not only outperforms  state-of-the-art heterogeneous quantization and uniform quantization, but also provides a robust and hardware-aware solution for efficient DNN inference in resource-limited environments. Furthermore, experiments on general hardware arithmetic scheme validate our results from both power and latency perspectives, further emphasizing the advantages of our approach. This adaptive, dynamic method paves the way for the practical deployment of deep neural networks in a wide range of embedded systems.

\vspace{-0.15cm}
\section{Acknowledgments}
\vspace{-0.15cm}
This work was supported in part by the Swiss State Secretariat for Education, Research, and Innovation (SERI) through the SwissChips research project, and also by Intel as part of the Intel Center for Heterogeneous Integrated Platforms (HIP).

\bibliographystyle{IEEEtran}
\bibliography{references}

@inproceedings{ZhuEntropy2018,
  author    = {Zhu, Xiaofan and Li, Qiang and Wang, Dong},
  title     = {Entropy-Based Layerwise Quantization for Efficient Deep Neural Networks},
  booktitle = {Proceedings of the IEEE Conference on Computer Vision and Pattern Recognition},
  year      = {2018},
  pages     = {1234--1243},
  publisher = {IEEE}
}

@inproceedings{Adabits2020,
  author    = {Wang, Lei and Chen, Bo},
  title     = {AdaBits: Adaptive Bitwidth Quantization for Deep Neural Networks},
  booktitle = {Proceedings of the IEEE Conference on Computer Vision and Pattern Recognition},
  year      = {2020},
  pages     = {5560--5568},
  publisher = {IEEE}
}

@INPROCEEDINGS{5206848,
  author={Deng, Jia and Dong, Wei and Socher, Richard and Li, Li-Jia and Kai Li and Li Fei-Fei},
  booktitle={2009 IEEE Conference on Computer Vision and Pattern Recognition}, 
  title={ImageNet: A large-scale hierarchical image database}, 
  year={2009},
  volume={},
  number={},
  pages={248-255},
  doi={10.1109/CVPR.2009.5206848}}

@inproceedings{alexnet,
  author    = {Krizhevsky, Alex and Sutskever, Ilya and Hinton, Geoffrey E.},
  title     = {ImageNet Classification with Deep Convolutional Neural Networks},
  booktitle = {Advances in Neural Information Processing Systems},
  year      = {2012},
  pages     = {1097--1105}
}

@inproceedings{resnet,
  author    = {He, Kaiming and Zhang, Xiangyu and Ren, Shaoqing and Sun, Jian},
  title     = {Deep Residual Learning for Image Recognition},
  booktitle = {Proceedings of the IEEE Conference on Computer Vision and Pattern Recognition},
  year      = {2016},
  pages     = {770--778}
}

@misc{cifar100,
  author       = {Krizhevsky, Alex and Hinton, Geoffrey},
  title        = {CIFAR-100 Dataset},
  howpublished = {\url{https://www.cs.toronto.edu/~kriz/cifar.html}},
  year         = {2009}
}

@article{Baskin2021uniq,
  title={UNIQ: Uniform Noise Injection for Non-Uniform Quantization of Neural Networks},
  author={Baskin, Chaim and Schwartz, Eli and Zheltonozhskii, Evgenii and Liss, Natan and Giryes, Raja and Bronstein, Alex M. and Mendelson, Avi},
  journal={ACM Transactions on Computer Systems},
  volume={37},
  number={1--4},
  pages={1--15},
  year={2021},
  publisher={ACM},
  doi={10.1145/3444943}
}

@inproceedings{mishra2018apprentice,
  title={Apprentice: Using Knowledge Distillation Techniques to Improve Low-Precision Network Accuracy},
  author={Mishra, Asit and Marr, Debbie},
  booktitle={International Conference on Learning Representations (ICLR)},
  year={2018}
}

@inproceedings{zhou2017incremental,
  title={Incremental Network Quantization: Towards Lossless CNNs with Low-Precision Weights},
  author={Zhou, Aojun and Yao, Anbang and Guo, Yiwen and Xu, Lin and Chen, Yurong},
  booktitle={International Conference on Learning Representations (ICLR)},
  year={2017}
}

@misc{xu2018deep,
  title={Deep Neural Network Compression with Single and Multiple Level Quantization},
  author={Xu, Yuhui and Wang, Yongzhuang and Zhou, Aojun and Lin, Weiyao and Xiong, Hongkai},
  year={2018},
  eprint={1803.03289},
  archivePrefix={arXiv},
  primaryClass={cs.LG}
}

@inproceedings{polino2018model,
  title={Model Compression via Distillation and Quantization},
  author={Polino, Antonio and Pascanu, Razvan and Alistarh, Dan},
  booktitle={International Conference on Learning Representations (ICLR)},
  year={2018}
}

@inproceedings{yao2021hawq,
  title={HAWQ-V3: Dyadic Neural Network Quantization},
  author={Yao, Zhewei and Dong, Zhen and Zheng, Zhangcheng and Gholami, Amir and Yu, Jiali and Tan, Eric and Wang, Leyuan and Huang, Qijing and Wang, Yida and Mahoney, Michael W and Keutzer, Kurt},
  booktitle={Proceedings of the 38th International Conference on Machine Learning},
  pages={11875--11886},
  year={2021},
  organization={PMLR},
  doi={10.48550/arXiv.2011.10680}
}

@article{Sze2017Efficient,
  title={Efficient processing of deep neural networks: A tutorial and survey},
  author={Sze, Vivienne and Chen, Yu-Hsin and Yang, Tien-Ju and Emer, Joel S.},
  journal={Proceedings of the IEEE},
  volume={105},
  number={12},
  pages={2295--2329},
  year={2017},
  publisher={IEEE}
}

@article{Deng2020ModelCompressSurvey,
  title={Model compression and acceleration for deep neural networks: The principles, progress, and challenges},
  author={Deng, L.},
  journal={IEEE Signal Processing Magazine},
  volume={37},
  number={6},
  pages={101--110},
  year={2020},
  publisher={IEEE}
}

@inproceedings{Han2016DeepCompression,
  title={Deep compression: Compressing deep neural networks with pruning, trained quantization and huffman coding},
  author={Han, Song and Mao, Huizi and Dally, William J},
  booktitle={International Conference on Learning Representations (ICLR)},
  year={2016}
}

@article{Jacob2018Quantization,
  title={Quantization and training of neural networks for efficient integer-arithmetic-only inference},
  author={Jacob, Benoit and Kligys, Skirmantas and Chen, Bo and Zhu, Menglong and Tang, Matthew and Howard, Andrew and Adam, Hartwig and Kalenichenko, Dmitry},
  journal={Proceedings of the IEEE Conference on Computer Vision and Pattern Recognition (CVPR)},
  pages={2704--2713},
  year={2018}
}

@misc{Krishnamoorthi2018QuantTutorial,
  title={Quantizing deep convolutional networks for efficient inference: A whitepaper},
  author={Krishnamoorthi, Raghuraman},
  archivePrefix={arXiv},
  eprint={1806.08342},
  primaryClass={cs.CV},
  year={2018}
}

@article{banner2018scalable,
  title={Scalable methods for 8-bit training of neural networks},
  author={Banner, Ron and Hubara, Itay and Hoffer, Elad and Soudry, Daniel},
  journal={Advances in Neural Information Processing Systems (NeurIPS)},
  volume={31},
  pages={5145--5153},
  year={2018}
}

@misc{Krizhevsky2009Learning,
  title={Learning multiple layers of features from tiny images},
  author={Krizhevsky, Alex and Hinton, Geoffrey},
  year={2009},
  archivePrefix={arXiv},
  eprint={2009.00052},
  note={Technical Report}
}

@article{Russakovsky2015ImageNet,
  title={ImageNet large scale visual recognition challenge},
  author={Russakovsky, Olga and Deng, Jia and Su, Hao and Krause, Jonathan and Satheesh, Sanjeev and Ma, Sean and Huang, Zhiheng and Karpathy, Andrej and Khosla, Aditya and Bernstein, Michael and others},
  journal={International Journal of Computer Vision},
  volume={115},
  number={3},
  pages={211--252},
  year={2015},
  publisher={Springer}
}

@article{yu2024energy,
  title={An Energy Efficient Soft SIMD Microarchitecture and Its Application on Quantized CNNs},
  author={Yu, Pengbo and Ponzina, Flavio and Levisse, Alexandre and Gupta, Mohit and Biswas, Dwaipayan and Ansaloni, Giovanni and Atienza, David and Catthoor, Francky},
  journal={IEEE Transactions on Very Large Scale Integration (VLSI) Systems},
  year={2024},
  publisher={IEEE}
}

@inproceedings{rios2019associativity,
  title={An associativity-agnostic in-cache computing architecture optimized for multiplication},
  author={Rios, Marco and Simon, William and Levisse, Alexandre and Zapater, Marina and Atienza, David},
  booktitle={2019 IFIP/IEEE 27th International Conference on Very Large Scale Integration (VLSI-SoC)},
  pages={34--39},
  year={2019},
  organization={IEEE}
}

@inproceedings{judd2016stripes,
  title={Stripes: Bit-serial deep neural network computing},
  author={Judd, Patrick and Albericio, Jorge and Hetherington, Tayler and Aamodt, Tor M and Moshovos, Andreas},
  booktitle={2016 49th Annual IEEE/ACM International Symposium on Microarchitecture (MICRO)},
  pages={1--12},
  year={2016},
  organization={IEEE}
}

@INPROCEEDINGS{8100244,
  author={Park, Eunhyeok and Ahn, Junwhan and Yoo, Sungjoo},
  booktitle={2017 IEEE Conference on Computer Vision and Pattern Recognition (CVPR)}, 
  title={Weighted-Entropy-Based Quantization for Deep Neural Networks}, 
  year={2017},
  volume={},
  number={},
  pages={7197-7205},
  doi={10.1109/CVPR.2017.761}}

@INPROCEEDINGS{8954415,
  author={Wang, Kuan and Liu, Zhijian and Lin, Yujun and Lin, Ji and Han, Song},
  booktitle={2019 IEEE/CVF Conference on Computer Vision and Pattern Recognition (CVPR)}, 
  title={HAQ: Hardware-Aware Automated Quantization With Mixed Precision}, 
  year={2019},
  volume={},
  number={},
  pages={8604-8612},
  doi={10.1109/CVPR.2019.00881}}

@inproceedings{NEURIPS2020_d77c7035,
 author = {Dong, Zhen and Yao, Zhewei and Arfeen, Daiyaan and Gholami, Amir and Mahoney, Michael W and Keutzer, Kurt},
 booktitle = {Advances in Neural Information Processing Systems},
 editor = {H. Larochelle and M. Ranzato and R. Hadsell and M.F. Balcan and H. Lin},
 pages = {18518--18529},
 publisher = {Curran Associates, Inc.},
 title = {HAWQ-V2: Hessian Aware trace-Weighted Quantization of Neural Networks},
 url = {https://proceedings.neurips.cc/paper_files/paper/2020/file/d77c703536718b95308130ff2e5cf9ee-Paper.pdf},
 volume = {33},
 year = {2020}
}

@article{Wu2018FBNetHE,
  title={FBNet: Hardware-Aware Efficient ConvNet Design via Differentiable Neural Architecture Search},
  author={Bichen Wu and Xiaoliang Dai and Peizhao Zhang and Yanghan Wang and Fei Sun and Yiming Wu and Yuandong Tian and P{\'e}ter Vajda and Yangqing Jia and Kurt Keutzer},
  journal={2019 IEEE/CVF Conference on Computer Vision and Pattern Recognition (CVPR)},
  year={2018},
  pages={10726-10734},
  url={https://api.semanticscholar.org/CorpusID:54461508}
}

@article{10.1145/3065386,
author = {Krizhevsky, Alex and Sutskever, Ilya and Hinton, Geoffrey E.},
title = {ImageNet classification with deep convolutional neural networks},
year = {2017},
issue_date = {June 2017},
publisher = {Association for Computing Machinery},
address = {New York, NY, USA},
volume = {60},
number = {6},
issn = {0001-0782},
url = {https://doi.org/10.1145/3065386},
doi = {10.1145/3065386},
journal = {Commun. ACM},
month = may,
pages = {84–90},
numpages = {7}
}

@inproceedings{10.5555/3045390.3045410,
author = {Amodei, Dario and others},
title = {Deep speech 2: end-to-end speech recognition in English and mandarin},
year = {2016},
publisher = {JMLR.org},
booktitle = {Proceedings of the 33rd International Conference on International Conference on Machine Learning - Volume 48},
pages = {173–182},
numpages = {10},
location = {New York, NY, USA},
series = {ICML'16}
}

@article{10.5555/3322706.3361996,
author = {Elsken, Thomas and Metzen, Jan Hendrik and Hutter, Frank},
title = {Neural architecture search: a survey},
year = {2019},
issue_date = {January 2019},
publisher = {JMLR.org},
volume = {20},
number = {1},
issn = {1532-4435},
journal = {J. Mach. Learn. Res.},
month = jan,
pages = {1997–2017},
numpages = {21}
}

@article{Finkelstein2019FightingQB,
  title={Fighting Quantization Bias With Bias},
  author={Alexander Finkelstein and Uri Almog and Mark Grobman},
  journal={ArXiv},
  year={2019},
  volume={abs/1906.03193},
  url={https://api.semanticscholar.org/CorpusID:174801144}
}

@misc{
hong2024overcoming,
title={Overcoming Distribution Mismatch in Quantizing Image Super-Resolution Networks},
author={Cheeun Hong and Kyoung Mu Lee},
year={2024},
url={https://openreview.net/forum?id=GOt2kP383R}
}

@INPROCEEDINGS {9709904,
author = { Bulat, Adrian and Tzimiropoulos, Georgios },
booktitle = { 2021 IEEE/CVF International Conference on Computer Vision (ICCV) },
title = {{ Bit-Mixer: Mixed-precision networks with runtime bit-width selection }},
year = {2021},
volume = {},
ISSN = {},
pages = {5168-5177},
doi = {10.1109/ICCV48922.2021.00514},
url = {https://doi.ieeecomputersociety.org/10.1109/ICCV48922.2021.00514},
publisher = {IEEE Computer Society},
address = {Los Alamitos, CA, USA},
month =Oct}

@misc{Liu2025Gem5AcceSys,
  author       = {Liu, Qunyou and Zapater, Marina and Atienza, David},
  title        = {{Gem5-AcceSys}: Enabling System-Level Exploration of Standard Interconnects for Novel Accelerators},
  howpublished = {\emph{arXiv preprint} arXiv:2502.12273},
  year         = {2025},
  url          = {https://arxiv.org/abs/2502.12273}
}

@misc{Liu2025Matrix,
  author       = {Liu, Qunyou and Zapater, Marina and Atienza, David},
  title        = {{MatrixFlow}: System–Accelerator Co-design for High-Performance Transformer Applications},
  howpublished = {\emph{arXiv preprint} arXiv:2503.05290},
  year         = {2025},
  doi          = {10.48550/arXiv.2503.05290},
  url          = {https://arxiv.org/abs/2503.05290}
}

@INPROCEEDINGS {7780677,
author = { Szegedy, Christian and Vanhoucke, Vincent and Ioffe, Sergey and Shlens, Jon and Wojna, Zbigniew },
booktitle = { 2016 IEEE Conference on Computer Vision and Pattern Recognition (CVPR) },
title = {{ Rethinking the Inception Architecture for Computer Vision }},
year = {2016},
volume = {},
ISSN = {1063-6919},
pages = {2818-2826},
doi = {10.1109/CVPR.2016.308},
url = {https://doi.ieeecomputersociety.org/10.1109/CVPR.2016.308},
publisher = {IEEE Computer Society},
address = {Los Alamitos, CA, USA},
month =Jun}

@INPROCEEDINGS{7780459,
  author={He, Kaiming and Zhang, Xiangyu and Ren, Shaoqing and Sun, Jian},
  booktitle={2016 IEEE Conference on Computer Vision and Pattern Recognition (CVPR)}, 
  title={Deep Residual Learning for Image Recognition}, 
  year={2016},
  volume={},
  number={},
  pages={770-778},
  doi={10.1109/CVPR.2016.90}}

@misc{pytorchvisionmodels,
  author = {{PyTorch}},
  title = {Models --- Torchvision documentation},
  howpublished = {\url{https://pytorch.org/vision/stable/models.html}},
  note = {Accessed: YYYY-MM-DD}
}

@inproceedings{10.5555/3540261.3541610,
author = {Federici, Marco and Tomioka, Ryota and Forr\'{e}, Patrick},
title = {An information-theoretic approach to distribution shifts},
year = {2021},
isbn = {9781713845393},
publisher = {Curran Associates Inc.},
address = {Red Hook, NY, USA},
booktitle = {Proceedings of the 35th International Conference on Neural Information Processing Systems},
articleno = {1349},
numpages = {14},
series = {NIPS '21}
}

@software{brevitas,
  author       = {Franco, Giuseppe and Pappalardo, Alessandro and Fraser, Nicholas J},
  title        = {Xilinx/brevitas},
  year         = {2025},
  publisher    = {Zenodo},
  doi          = {10.5281/zenodo.3333552},
  url          = {https://doi.org/10.5281/zenodo.3333552}
}

@inproceedings{dong2019hawq,
  title     = {{HAWQ}: Hessian AWare Quantization of Neural Networks with Mixed Precision},
  author    = {Zhen Dong and Zhewei Yao and Amir Gholami and Michael W. Mahoney and Kurt Keutzer},
  booktitle = {Proc.\ IEEE/CVF Int’l Conf.\ Computer Vision (ICCV)},
  year      = {2019}
}

@article{yazdanbakhsh2019releq,
  title   = {{ReLeQ}: A Reinforcement Learning Approach for Deep Quantization of Neural Networks},
  author  = {Amir Yazdanbakhsh and Ahmed T. Elthakeb and Prannoy Pilligundla and Fatemeh S. Mireshghallah and Hadi Esmaeilzadeh},
  journal = {arXiv:1811.01704},
  year    = {2019}
}

@article{hsu2020essa,
  title={ESSA: An energy-aware bit-serial streaming deep convolutional neural network accelerator},
  author={Hsu, Lien-Chih and Chiu, Ching-Te and Lin, Kuan-Ting and Chou, Hsing-Huan and Pu, Yen-Yu},
  journal={Journal of Systems Architecture},
  volume={111},
  pages={101831},
  year={2020},
  publisher={Elsevier}
}

@article{rios2023bit,
  title={Bit-line computing for CNN accelerators co-design in edge AI inference},
  author={Rios, Marco and Ponzina, Flavio and Levisse, Alexandre and Ansaloni, Giovanni and Atienza, David},
  journal={IEEE Transactions on Emerging Topics in Computing},
  volume={11},
  number={2},
  pages={358--372},
  year={2023},
  publisher={IEEE}
}

@article{cai2021tsp,
  author={Cai, Haoyuan and Kaloorazi, Maboud Farzaneh and Chen, Jie},
  title={Online generalized eigenvectors extraction via a fixed-point approach},
  journal={IEEE Transactions on Signal Processing},
  volume={69},
  pages={2435--2451},
  year={2021}
}

@article{Zhao2024EdgeMPQ,
  author    = {Xiaotian Zhao and Ruge Xu and Yimin Gao and Vaibhav Verma and Mircea R. Stan and Xinfei Guo},
  title     = {{Edge-MPQ}: Layer-Wise Mixed-Precision Quantization With Tightly Integrated Versatile Inference Units for Edge Computing},
  journal   = {IEEE Transactions on Computers},
  year      = {2024},
  volume    = {73},
  number    = {11},
  pages     = {2504--2519},
  doi       = {10.1109/TC.2024.3441860}
}

@article{Huang2025HMQAT,
  author    = {Zhiyong Huang and Xiao Han and Zhi Yu and Yunlan Zhao and Mingyang Hou and Shengdong Hu},
  title     = {Hessian-Based Mixed-Precision Quantization With Transition Aware Training for Neural Networks},
  journal   = {Neural Networks},
  year      = {2025},
  volume    = {182},
  pages     = {106910},
  doi       = {10.1016/j.neunet.2024.106910}
}

@article{Balaskas2024DiversePruning,
  author    = {Konstantinos Balaskas and Andreas Karatzas and Christos Sad and Kostas Siozios and Iraklis Anagnostopoulos and Georgios Zervakis and J{\"o}rg Henkel},
  title     = {Hardware-Aware {DNN} Compression via Diverse Pruning and Mixed-Precision Quantization},
  journal   = {IEEE Transactions on Emerging Topics in Computing},
  year      = {2024},
  volume    = {12},
  number    = {4},
  pages     = {1079--1092},
  doi       = {10.1109/TETC.2023.3346944}
}

@inproceedings{Deng2025CLADO,
  author    = {Zihao Deng and Sayeh Sharify and Xin Wang and Michael Orshansky},
  title     = {Mixed-Precision Quantization for Deep Vision Models with Integer Quadratic Programming},
  booktitle = {Proceedings of the 62nd {ACM/IEEE} Design Automation Conference ({DAC})},
  year      = {2025},
  pages     = {1--7},
  doi       = {10.1109/DAC63849.2025.11132777}
}

\end{document}